\documentclass[10pt,journal,compsoc]{IEEEtran}
\usepackage[nocompress]{cite}

\usepackage[margin=0pt,font=small,labelfont=bf,labelsep=endash,tableposition=top]{caption}

\usepackage{array}
\usepackage[caption=false,font=footnotesize,labelfont=sf,textfont=sf]{subfig}
\usepackage{url}
\usepackage{booktabs}
\usepackage{dsfont}
\usepackage{pifont}
\usepackage{colortbl}
\usepackage{graphicx}
\usepackage{amsmath}
\usepackage{amssymb}
\usepackage{epsfig}
\usepackage{epstopdf}
\usepackage{makecell,multirow,diagbox}
\usepackage{color}
\usepackage{soul}
\usepackage{url}
\usepackage{amsmath}
\usepackage[utf8]{inputenc}
\usepackage{xcolor}
\usepackage{mathtools,xspace}
\usepackage{ragged2e}
\usepackage{bm}
\usepackage[pagebackref,breaklinks,colorlinks]{hyperref}

\newcommand{\onedot}{\ifx\@let@token.\else.\null\fi\xspace}
\newcommand{\etal}{\emph{et al}\onedot}
\newcommand{\eg}{\emph{e.g}\onedot}
\newcommand{\ie}{\emph{i.e}\onedot}
\newcommand{\etc}{\emph{etc}}

\newcommand{\xmark}{\textcolor{red}{\ding{56}}}
\newcommand{\tmark}{\textcolor[rgb]{0,0.8,0}{\ding{52}}}

\begin{document}

\title{Hi-SAM: Marrying Segment Anything Model for Hierarchical Text Segmentation}

\author{
Maoyuan Ye,
Jing Zhang,~\IEEEmembership{Senior Member,~IEEE},
Juhua Liu,~\IEEEmembership{Member,~IEEE,}
Chenyu Liu,
Baocai Yin,\\
Cong Liu,
Bo Du,~\IEEEmembership{Senior Member,~IEEE},
Dacheng Tao,~\IEEEmembership{Fellow,~IEEE}

\thanks{This work was supported in part by the National Key Research and Development Program of China under Grant 2023YFC2705700 and 2022YFB4500600, in part by the National Natural Science Foundation of China under Grants U23B2048, 62076186 and 62225113, in part by the Innovative Research Group Project of Hubei Province under Grant 2024AFA017, and in part by the Science and Technology Major Project of Hubei Province under Grant 2024BAB046. The numerical calculations in this paper have been done on the supercomputing system in the Supercomputing Center of Wuhan University. Corresponding authors: Juhua Liu, Bo Du (e-mail: \{liujuhua, dubo\}@whu.edu.cn).}

\thanks{M. Ye, J. Zhang, J. Liu, and B. Du are with the School of Computer Science, National Engineering Research Center for Multimedia Software, Institute of Artificial Intelligence, and Hubei Key Laboratory of Multimedia and Network Communication Engineering, Wuhan University, Wuhan, China (e-mail: \{yemaoyuan, liujuhua, dubo\}@whu.edu.cn, jingzhang.cv@gmail.com).}

\thanks{C. Liu, B. Yin and C. Liu are with the iFLYTEK Research, IFLYTEK CO. LTD., China (email: \{cyliu7, bcyin, congliu2\}@iflytek.com).} 

\thanks{D. Tao is with the College of Computing \& Data Science at Nanyang Technological University, \#32 Block N4 \#02a-014, 50 Nanyang Avenue, Singapore 639798 (e-mail: dacheng.tao@ntu.edu.sg).}
}

\IEEEtitleabstractindextext{%
\begin{abstract}
\justifying
The Segment Anything Model (SAM), a profound vision foundation model pretrained on a large-scale dataset, breaks the boundaries of general segmentation and sparks various downstream applications. This paper introduces \textbf{Hi-SAM}, a unified model leveraging SAM for hierarchical text segmentation. Hi-SAM excels in segmentation across four hierarchies, including \textit{pixel-level text}, \textit{word}, \textit{text-line}, and \textit{paragraph}, while realizing \textit{layout analysis} as well. Specifically, we first turn SAM into a high-quality pixel-level text segmentation (TS) model through a parameter-efficient fine-tuning approach. We use this TS model to iteratively generate the pixel-level text labels in a semi-automatical manner, unifying labels across the four text hierarchies in the HierText dataset. Subsequently, with these complete labels, we launch the end-to-end trainable Hi-SAM based on the TS architecture with a customized hierarchical mask decoder. 
During inference, Hi-SAM offers both automatic mask generation (AMG) mode and promptable segmentation (PS) mode. In the AMG mode, Hi-SAM segments pixel-level text foreground masks initially, then samples foreground points for hierarchical text mask generation and achieves layout analysis in passing. As for the PS mode, Hi-SAM provides word, text-line, and paragraph masks with a single point click.
Experimental results show the state-of-the-art performance of our TS model: 84.86\% fgIOU on Total-Text and 88.96\% fgIOU on TextSeg for pixel-level text segmentation. Moreover, compared to the previous specialist for joint hierarchical detection and layout analysis on HierText, Hi-SAM achieves significant improvements: 4.73\% PQ and 5.39\% F1 on the text-line level, 5.49\% PQ and 7.39\% F1 on the paragraph level layout analysis, requiring $20\times$ fewer training epochs. 
The code is available at \href{https://github.com/ymy-k/Hi-SAM}{Hi-SAM}.

\end{abstract}

}

\maketitle

\IEEEdisplaynontitleabstractindextext
\IEEEpeerreviewmaketitle

\IEEEraisesectionheading{\section{Introduction}
\label{sec:introduction}}
\IEEEPARstart{T}{ext} effectively conveys rich high-level semantics through various hierarchies, spanning from pixel-level text to word, text-line, and paragraph. These hierarchies play crucial roles in diverse applications. For instance, pixel-level text segmentation (TS)\footnote{We view \textit{text segmentation} in previous works~\cite{xu2021rethinking,yu2023scene,peng2023upocr} and \textit{pixel-level text segmentation} as the same task. To avoid confusion with other text hierarchies, we always use \textit{pixel-level text segmentation} in this paper.}~\cite{xu2021rethinking,yu2023scene,peng2023upocr} is useful for font style transfer, scene text removal, and text editing. Additionally, word and text-line detection are important for text entity extraction and recognition, while visual text understanding~\cite{appalaraju2021docformer} may necessitate a geometric layout, such as paragraph grouping results~\cite{long2022towards,long2024hierarchical}. Existing works~\cite{liu2020ASTS,xu2021rethinking,ren2022looking,xu2022bts,zu2023weakly,liao2020real,zhang2022arbitrary,ye2023dptext,ye2023deepsolo,kittenplon2022towards,liu2023polyformer} typically concentrate on specific hierarchies, lacking a comprehensive approach to handle multi-grained textual information within a unified system. This raises the question: \textbf{\textit{
Is it feasible to devise a cohesive segmentation framework capable of effectively managing diverse text hierarchies, encompassing pixel-level text, word, text-line, and paragraph?
}}
A major challenge lies in the limited availability of real-world data annotated with all four text hierarchies. Table~\ref{tab1:inconsistency} briefly summarizes existing datasets, revealing a predominant focus on word level. Only HierText~\cite{long2022towards} provides hierarchical annotations of word, text-line, and paragraph, yet it lacks pixel-level text annotations and the training set volume is also limited.
This suggests the need for data annotation of all required text hierarchies as well as data-efficient training in hierarchical text segmentation.

\begin{table}[t]
    \centering
    \setlength{\tabcolsep}{1pt}
    \caption{
    \textbf{Current datasets lack comprehensive and cohesive text hierarchies}. Our study fills this gap by introducing pixel-level text labels to HierText, establishing it as the pioneering dataset with four distinct text hierarchies. The statistics of words are achieved from training and validation sets.
    }
    \label{tab1:inconsistency}
    \resizebox{\hsize}{!}{
    \begin{tabular}{ccccccc}
         \toprule[1.1pt]
         \rowcolor{gray!20} & &Words & \multicolumn{4}{c}{Hierarchies} \\
         \cline{4-7}
         \rowcolor{gray!20}\multirow{-2}{*}{Datasets} &\multirow{-2}{*}{Training Images} &(avg/total) &Pixel-level Text &Word &Line &Paragraph \\
         \midrule[1pt]
         MSRA-TD500~\cite{yao2012detecting} &300 &6.9/3.5K &\xmark &\xmark &\tmark &\xmark \\
         IC15~\cite{karatzas2015icdar} &1,000 &4.4/6.5K &\xmark &\tmark &\xmark &\xmark \\
         CTW1500~\cite{liu2019curved} &1,000 &6.7/10K &\xmark &\xmark &\tmark &\xmark \\
         Total-Text~\cite{ch2017total} &1,255 &7.4/11K &\tmark &\tmark &\xmark &\xmark \\
         TextSeg~\cite{xu2021rethinking} &2,646 &3.9/12K &\tmark &\tmark &\xmark &\xmark \\
         IC19-ArT~\cite{chng2019icdar2019} &5,603 & 8.9/50K &\xmark &\tmark &\xmark &\xmark \\
         IC19-LSVT~\cite{sun2019icdar} &30,000 & 8.1/243K &\xmark &\xmark &\tmark &\xmark \\
         TextOCR~\cite{singh2021textocr} &21,778 &32.1/903K &\xmark &\tmark &\xmark &\xmark \\
         HierText~\cite{long2022towards} &8,281 &103.8/1.2M &\xmark &\tmark &\tmark &\tmark \\
         \midrule[1pt]
         HierText* (w/ SAM-TS) &8,281 &103.8/1.2M &\tmark &\tmark &\tmark &\tmark \\
         \bottomrule[1.1pt]
    \end{tabular}}
    \vspace{-2mm}
\end{table}

%---------------------------------
\begin{figure*}[!t]
    \centering
    \includegraphics[width=1\linewidth]{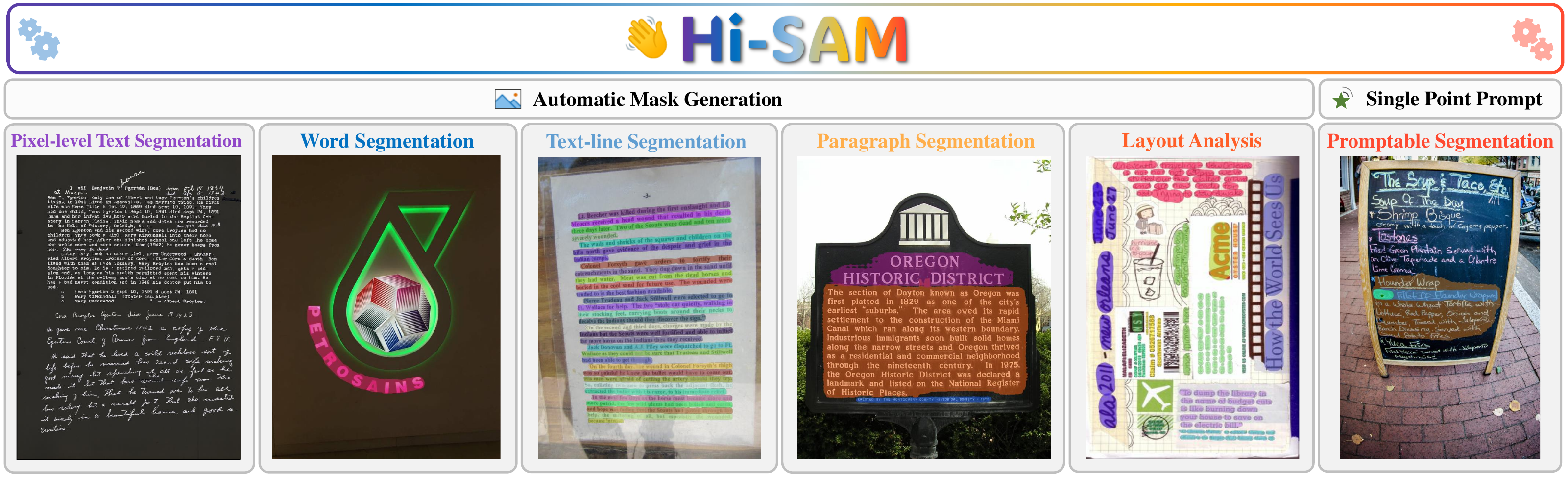}
    \caption{Hi-SAM can perform pixel-level text segmentation, word segmentation, text-line segmentation, paragraph segmentation, and layout analysis in automatic mask generation mode. Hi-SAM also supports promptable segmentation. Given a single-point click on one word, Hi-SAM predicts the corresponding word, text-line, and paragraph masks.}
    \label{fig1}
\end{figure*}
%---------------------------------

Recently, the Segment Anything Model (SAM)~\cite{kirillov2023segment} emerged as a foundational vision model for general image segmentation, leveraging billion-scale class-agnostic mask labels for training. SAM exhibits promptable and multi-grained segmentation capabilities, generating object masks based on point, bounding box, or coarse mask inputs. Notably, SAM demonstrates robust transferability in zero-shot scenarios and task adaptation across diverse domains, including common image segmentation~\cite{kirillov2023segment,sam_hq}, medical image analysis~\cite{cheng2023sam,yue2023surgicalsam}, and video processing~\cite{cheng2023segment,yang2023track}. Additionally, SAM facilitates scalable data annotation~\cite{SAMRS}. 
Motivated by SAM's excellence, we aim to answer the above question by harnessing its capabilities to build a strong hierarchical text segmentation model.

To this end, we propose Hi-SAM, a text-centric hierarchical segmentation model that leverages the inherent knowledge of SAM as well as existing fragmented training data with incomplete text hierarchy annotations.
In the automatic mask generation (AMG) mode, SAM employs class and object-agnostic grid points as prompts for segmenting objects, ranging from groups to individual parts, based on the best confidence across scales. 
However, for hierarchical text segmentation, the text must be considered as the sole foreground and it is required to explicitly and simultaneously generate word, text-line, and paragraph masks. To address this issue, we propose to segment the pixel-level text masks at first, and then sample points from them as prompts for generating word, text-line, and paragraph masks subsequently. 
For the first step, we surprisingly discovered that SAM can function as a competitive TS model with minimal customizations.
Unlike SAM's original mask decoder which takes local point or box-based prompts for segmentation, we propose to use global prompts, enabling the mask decoder to perceive the overall context of image texts. Technically, we convert the image encoder features to a sequence of tokens and use them as implicit prompts to tune the mask decoder. Furthermore, we identify the bottleneck in SAM's mask decoder for segmenting text with fine details as the small mask feature size. Consequently, we introduce a simple yet effective module to empower the mask decoder in delivering text features at the original input size. 
We call this simple baseline model SAM-TS and demonstrate that it surpasses state-of-the-art (SOTA) TS models across various existing datasets. 
In addition, inspired by SAM, we adopt an iterative strategy to first train SAM-TS on HierText~\cite{long2022towards} using a small set of manually labeled data, and then annotate additional pixel-level text in HierText by leveraging the trained SAM-TS with minimal manual intervention.

Next, we launch Hi-SAM, an end-to-end trainable model for hierarchical text segmentation, by extending SAM-TS. Hi-SAM incorporates a hierarchical mask decoder (H-Decoder) parallel to SAM-TS's mask decoder (S-Decoder), enabling segmentation at word, text-line, and paragraph levels. Foreground points are randomly sampled from the predicted pixel-level text mask and transformed into point prompts using SAM's prompt encoder. For each point prompt that belongs to a certain text, the H-Decoder uses three output tokens to predict its hierarchical masks. Hi-SAM supports both the AMG mode and promptable segmentation (PS) mode as shown in Fig.~\ref{fig1}. In the AMG mode, Hi-SAM initially segments the pixel-level text mask, samples point prompts, and generates hierarchical masks. In the PS mode, a single-point click on a word prompts Hi-SAM to generate the word mask, along with text-line and paragraph masks at the word's location. Additionally, Hi-SAM provides layout analysis as a by-product by calculating Intersection-over-Union (IoU) matrix of paragraph masks and using it for clustering words.

In summary, our main contributions are three-fold:
\begin{itemize}
    \item We propose a simple yet effective method to render SAM as a competitive TS baseline model, \textit{i.e.}, SAM-TS. It surpasses SOTA TS methods, producing high-resolution masks logits. We further employ it to semi-automatically annotate pixel-level text labels for HierText, making it the first real dataset featuring mask annotations across four text hierarchies.
    \item We introduce Hi-SAM that builds upon SAM-TS with simple add-ons and is trained end-to-end on the enhanced HierText. Hi-SAM is the first method for hierarchical text segmentation, covering pixel-level text, word, text-line, and paragraph levels, and supporting both AMG and PS modes.
    \item Our SAM-TS and Hi-SAM offer compelling performance on public and challenging datasets. SAM-TS sets new records on Total-Text (84.86\% fgIOU ) and TextSeg (88.96\% fgIOU). Hi-SAM surpasses previous specialists for joint text detection and layout analysis on HierText by a large margin while requiring 20$\times$ fewer training epochs.
\end{itemize}

The paper is structured as follows: Section \ref{sec:related work} provides a brief review of related works. Section \ref{sec:method} introduces the proposed method. Extensive experimental results are presented in Section \ref{sec:experiments}, followed by a discussion on Hi-SAM's limitations in Section \ref{sec:limitation_discussion}. The paper concludes in Section \ref{sec:conclusion}.

\section{Related Work}
\label{sec:related work}
\subsection{Specialist Models for Different Text Hierarchies}
\textbf{(i) Pixel-level Text.} Pixel-level text segmentation requires segmenting all text strokes from the background. It presents a fine-grained per-pixel binary segmentation challenge. The pixel of text is considered as the foreground. Existing datasets primarily target scene text and designed text of large or medium sizes, such as Total-Text~\cite{ch2017total} and TextSeg~\cite{xu2021rethinking}. To address the limited training data, some methods integrate text recognizers to enhance multi-scale features~\cite{ren2022looking, wang2023textformer, xu2022bts}. Additionally, the recent approach~\cite{zu2023weakly} adopts a weakly supervised approach by combining pixel-level text segmentation with text recognition. In contrast to prior studies, our work demonstrates the utilization of SAM's knowledge, leveraging single-scale features from the ViT encoder to achieve SOTA performance.
\textbf{(ii) Word and Text-line.} 
Existing models predominantly focus on word-level scene text detection, with early approaches~\cite{liao2017textboxes,liao2018textboxes++} employing axis-aligned or multi-oriented anchor boxes for word localization. To address arbitrarily-shaped text representation, some methods~\cite{baek2019character,wang2019shape,wang2019efficient,ye2020textfusenet,liao2020real,zhang2021adaptive} utilize pixel-level segmentation masks. For example, DB~\cite{liao2020real} segments text kernel regions to distinguish instances, performs an adaptive binarization process, and employs post-processing to expand kernel contours for final results. Alternatively, some methods regress control points~\cite{dai2021progressive,zhang2022text,ye2023dptext} or predict parameterized curves~\cite{liu2020abcnet,zhu2021fourier} to fit text contours. However, these approaches often require two sets of weights for word and text-line detection, lacking a unified model.
\textbf{(iii) Layout Analysis and Document Understanding.} Layout analysis primarily focuses on document scenarios, treating visually and semantically cohesive text blocks as either detection~\cite{schreiber2017deepdesrt} or segmentation~\cite{lee2019page} objects. 
In addition to text, some document datasets~\cite{zhong2019publaynet} require distinguishing different elements, including title, list, table, and figure. Biswas \etal~\cite{biswas2021beyond} establish a Mask-RCNN~\cite{he2017mask} based model for complex document layouts. DocSegTr~\cite{biswas2022docsegtr} is presented as a Transformer-based model. Language-model-based approaches~\cite{huang2022layoutlmv3,li2021structurallm} leverage Optical Character Recognition (OCR) tokens to group words into segments for semantic parsing. Recently, Long \etal~\cite{long2022towards} introduce a Unified Detector (UD) that addresses joint text detection and layout analysis in both natural and document scenarios. Since the elements like table scarcely exist in natural images, text is considered as the sole object. In UD, each object query is trained to segment words within a text line, and a layout branch generates an affinity matrix between different object queries for clustering into paragraphs. 
On the other hand, there are several segmentation-free methods for document recognition and understanding. For example, Donut~\cite{kim2022ocr}, an OCR-free Transformer-based model, predicts a sequence of tokens that can be converted into target information in a structured form to realize visual document understanding. 
Similarly, Dessurt~\cite{davis2022end} also achieves document understanding with Transformer in the autoregressive text generation manner. 
DAN~\cite{coquenet2023dan} is the first architecture for handwritten document recognition that is able to recognize text at document level while labeling logical layout information. DAN sequentially outputs characters, as well as logical layout tokens, labeling text parts with begin and end tags in an XML-like fashion. 
Pix2Struct~\cite{lee2023pix2struct} is proposed with a screenshot parsing pretraining objective based on the HTML format of web pages, enabling more effective pixel-to-text transformation for general-purpose visually-situated language understanding.
To summarize, for high-level visual-language understanding tasks, such as visual question answering and information extraction, segmentation is not a necessity. However, recognition and visual-language understanding are not the focus of this work. In contrary, we concentrate on the pixel-level OCR tasks and concern the multi-granularity problem~\cite{kirillov2023segment,li2023semantic} which is crucial in computer vision community.

Although text naturally exhibits multi-granularity across various hierarchies from pixel-level text, to word, text-line, and paragraph, prior to this research, a unified segmentation framework had not been explored. Previous studies have predominantly focused on specific text hierarchies. Their differing paradigms, architectures, and training strategies add complexity, raising both deployment and maintenance costs. In this work, we aim to explore the first unified framework for these text hierarchies, enabling one model to automatically complete these tasks. Furthermore, there is currently no text segmentation model that supports prompts. A promptable model could assist in annotation and interactive applications. For instance, with a single prompt, the model can segment an object according to designated text hierarchies, thereby reducing annotation time. Moreover, it can facilitate interactive pixel-level image editing at specified locations and text hierarchies. Hence, we investigate this automatic and promptable model, which opens up new application possibilities.

\subsection{Adapting Vision Foundation Model for Text Tasks}
Recent studies~\cite{song2022vision,xue2022language,yu2023turning,yu2023turningspotter} have delved into leveraging CLIP~\cite{radford2021learning} to improve backbone representations for scene text detection and spotting. Specifically, oCLIP~\cite{xue2022language} introduces a weakly supervised pretrained network aligning visual and partial textual information for scene text detection and spotting. In contrast, TCM~\cite{yu2023turning} adapts the CLIP model for scene text detection through visual prompt tuning. Additionally, for enhanced scene text recognition, CLIP-OCR~\cite{wang2023symmetrical} proposes a symmetrical distillation strategy capturing linguistic knowledge in the CLIP text encoder. In our work, we advance this field by developing a hierarchical and promptable segmentation framework using SAM~\cite{kirillov2023segment}.

%---------------------------------
\begin{figure*}[!t]
    \centering
    \includegraphics[width=1\linewidth]{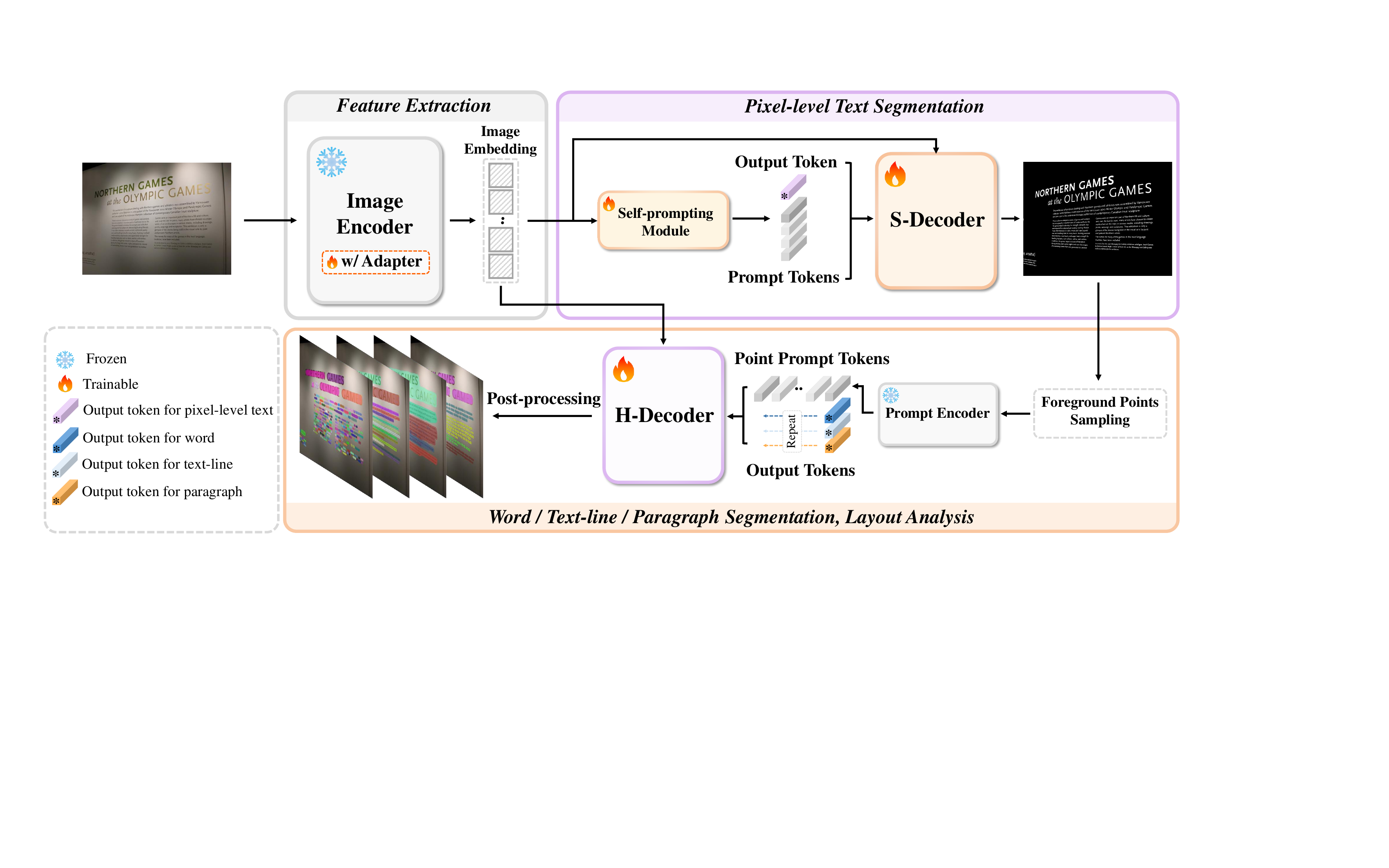}
    \caption{\textbf{The overview of Hi-SAM.} We show the automatic mask generation mode here. With the image embedding, for pixel-level text segmentation, self-prompting module generates implicit prompt tokens for the mask decoder (S-Decoder). Based on the pixel-level text mask, a certain number of foreground points are sampled and then embedded by the frozen prompt encoder. A customized hierarchical mask decoder (H-Decoder) segments word, text-line, and paragraph masks for each point prompt. Layout analysis can be achieved with the hierarchical outputs from the H-Decoder in passing.
    }
    \label{fig: hi-sam structure}
    \vspace{-2mm}
\end{figure*}
%---------------------------------

\subsection{Segment Anything Model and Follow-ups}
SAM~\cite{kirillov2023segment} is a pioneering vision model for general image segmentation, exhibiting remarkable generalization capabilities through large-scale pretraining. It extends its impact to diverse downstream tasks such as image matting~\cite{li2024matting}, 3D segmentation~\cite{cen2023segment}, video tracking~\cite{cheng2023segment,yang2023track}, and medical image segmentation~\cite{yue2023surgicalsam,yue2023part}. For example, SAM faces challenges in medical image segmentation due to domain gaps. Some methods~\cite{wu2023medical,cheng2023sam} leverage adapter tuning on SAM's ViT encoder to better adapt it to the medical imaging domain. PerSAM~\cite{zhang2023personalize} and Matcher~\cite{liu2023matcher} introduce training-free segmentation frameworks based on SAM with one-shot learning. HQ-SAM~\cite{sam_hq} enhances SAM's object segmentation quality while maintaining zero-shot generalizability by applying a learnable high-quality output token on fused features of size $256\times256$ to improve mask details. Despite extensive studies in various domains, there is a notable gap in research on text-centric segmentation. In this work, we introduce the first unified segmentation framework for four text hierarchies. We identify the primary limitation of applying SAM to fine-grained text segmentation in the mask feature size. Our approach employs a simple yet effective method to achieve high-quality pixel-level text segmentation by providing high-resolution mask features.

\section{Methodology}
\label{sec:method}
In this work, we propose Hi-SAM which excels in unified text segmentation spanning four hierarchies, including pixel-level text, word, text-line, and paragraph, while realizing layout analysis in passing. In the following subsections, we first briefly review the preliminaries about SAM in Sec.~\ref{subsec: reviewing sam}. Then, we provide a detailed description of Hi-SAM.

\subsection{Preliminary}
\label{subsec: reviewing sam}
SAM~\cite{kirillov2023segment} consists of a ViT backbone, a prompt encoder, and a lightweight mask decoder. The backbone (ViT-B/ViT-L/ViT-H) extracts image embeddings of size $64\times64$. SAM initializes the ViT with MAE~\cite{he2022masked} pretrained weights before training. The prompt encoder embeds different types of interactive positional cues into prompt tokens. The two-way Transformer-based mask decoder takes image embeddings, output tokens, and prompt tokens as inputs. The output tokens are similar to the object queries in DETR~\cite{carion2020end}. The mask decoder uses a multi-layer perceptron (MLP) to predict dynamic weights based on the output tokens, then applies the dynamic weights on up-sampled mask features in $256\times256$ spatial shape to generate mask results.
Although SAM shows impressive generalization capabilities, it remains a class-agnostic model. How to leverage SAM for text-centric hierarchical segmentation is a pending issue.

\subsection{Overview of Hi-SAM}
\label{subsec: hi-sam}
Hi-SAM employs a unified framework for hierarchical text segmentation, initially tackling the global TS task and subsequently addressing local segmentation tasks. The overall architecture is depicted in Fig.~\ref{fig: hi-sam structure}. Hi-SAM contains five modules: 1) an image encoder from SAM, 2) a plug-in self-prompting module, 3) a pixel-level text mask decoder (S-Decoder) derived from SAM's mask decoder, 4) a frozen prompt encoder from SAM, and 5) a hierarchical mask decoder (H-Decoder) customized from SAM's mask decoder.
For TS, we use the self-prompting module and S-Decoder to process the image embedding. Recognizing the fine details inherent in pixel-level text, we also devise a simple yet effective module to generate high-resolution pixel-level text mask features, thereby significantly enhancing performance. This base component of Hi-SAM for TS, \ie, including the image encoder, self-prompting module, and S-Decoder, is referred to as SAM-TS. It is used to generate the labels of pixel-level text for HierText in a semi-automatic way. The details of SAM-TS are presented in Sec.~\ref{subsubsec: text segmentation}.
After obtaining the masks of pixel-level text, a certain number of foreground points are sampled as the bridge to word, text-line, and paragraph segmentation. With these foreground points, the prompt encoder and H-Decoder are used to predict masks for the remaining three text hierarchies. This setting is referred to as the AMG mode while Hi-SAM also retains the PS mode of SAM. We introduce the details of the H-decoder in Sec.~\ref{subsubsec: wtp segmentation}. 
Given the paragraph masks, Hi-SAM is capable of conducting layout analysis, which is described in Sec.~\ref{subsubsec: layout}. We detail the training and inference process of Hi-SAM in Sec.~\ref{subsubsec: training} and Sec.~\ref{subsubsec: inference}.

\subsection{Feature Extraction}
Since SAM's image encoder is not fully aware of text with fine details, directly applying the frozen image encoder for pixel-level text segmentation cannot achieve satisfactory results. To efficiently enhance the adaptability of the image encoder, an adapter-tuning approach~\cite{wu2023medical} is adopted but is not constrained as the sole choice. The original parameters of SAM's image encoder are maintained frozen while two trainable adapter modules are inserted in each ViT block. Each adapter consists of a down-projection, a ReLU activation, and an up-projection sequentially. We use the default setting in~\cite{wu2023medical}.
Next, we devise a self-prompting module to generate implicit prompt tokens from the image embedding and turn SAM's mask decoder into the S-Decoder with high-resolution mask features.

\subsection{Pixel-level Text Segmentation}
\label{subsubsec: text segmentation}
\noindent\textbf{Self-prompting Module.} SAM's mask decoder takes prompt tokens with explicit positional cues as inputs. To steer the S-Decoder to segment pixel-level text within the input image, we propose to use the image embedding itself to generate implicit prompt tokens. 
The idea is that the required prompts for pixel-level text are already stored in image embedding, we just need to convert them into a series of representative prompt tokens.
To achieve this, in the self-prompting module, we start by formulating an image tokenizer function $T$: 
\begin{equation}
\label{eq: tokenizer}
    \bm{t} = T(\bm{I}),
\end{equation}
which converts the image embedding $\bm{I} \in \mathbb{R}^{64 \times 64 \times 256}$ into tokens $\bm{t} \in \mathbb{R}^{N \times 256}$. $64 \times 64$ denotes the spatial shape of image embedding after the ViT backbone, $256$ is the feature dimension. $N$ is the token length.
Inspired by TokenLearner~\cite{ryoo2021tokenlearner}, we use a convolution block $ConvBlock$ and a $sigmoid$ function to process $\bm{I}$, generating a spatial attention map $\bm{A} \in \mathbb{R}^{64 \times 64 \times N}$:
\begin{equation}
    \bm{A} = sigmoid(ConvBlock(\bm{I})),
\end{equation}
where $ConvBlock$ contains four convolution layers (kernel size $3 \times 3$, stride 1, padding 1) with the first one decreasing the dimension from $256$ to $N$. GELU activation is inserted between convolution layers.
Then we use $\bm{A}$, $\bm{I}$, and the spatial global average pooling operation $\rho$ to obtain the tokens. We rewrite Eq. (\ref{eq: tokenizer}) as following:
\begin{equation}
    \bm{t} = T(\bm{I}) = \rho(\gamma_{A}(\bm{A}) \odot \gamma_{I}(\bm{I})),
\end{equation}
where $\odot$ is the element-wise product, $\gamma_{A}$ and $\gamma_{I}$ stand for reshape and broadcast operations on spatial attention map $\bm{A}$ and image embedding $\bm{I}$. 
To be more specific, for each channel of the spatial attention map $\bm{A}$, it is broadcasted to $256$ dimension and multiplied with $\bm{I}$, achieving the spatially focused image embedding $\bm{I}' \in \mathbb{R}^{64 \times 64 \times 256}$. Then, a token with the shape of $1 \times 256$ can be calculated with the pooling operation $\rho$ on $\bm{I}'$. In this way, with $\bm{A}$ owning $N$ channels, the tokens $\bm{t} \in \mathbb{R}^{N \times 256}$ can be obtained.
The tokens $\bm{t}$ are further sent into a single Transformer decoder layer $TDecLayer$ for enhancing their representations:
\begin{equation}
    \bm{t_{prompt}} = TDecLayer(Q=\bm{t}, K=\bm{I}, V=\bm{I}),
\end{equation}
where $\bm{t_{prompt}} \in \mathbb{R}^{N \times 256}$ denotes the final prompt tokens that will be fed into the S-Decoder. $TDecLayer$ takes $\bm{t}$ as query, $\bm{I}$ as key and value.

\noindent\textbf{S-Decoder with High-resolution Mask Features.}
We observe that the mask feature resolution mainly constrains the pixel-level text segmentation quality. Therefore, we design a simple yet effective method to provide high-resolution mask feature, significantly boosting the performance of pixel-level text with fine structures (See ablation results in Tab.~\ref{tab: effect of HR}).

\begin{figure}[!t]
    \centering
    \includegraphics[width=1\linewidth]{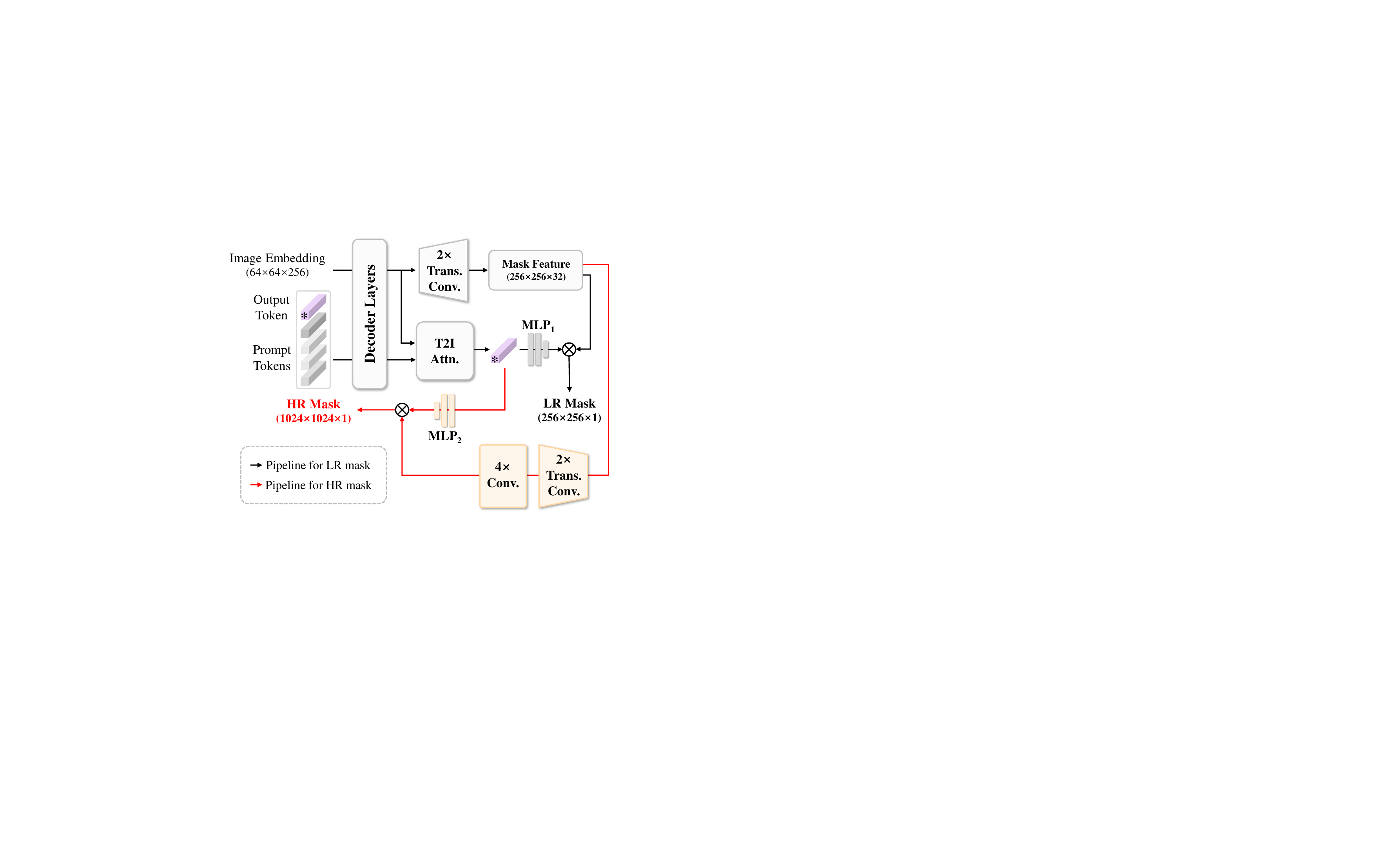}
    \caption{\textbf{The structure details of S-Decoder.} `Trans. Conv.' and `T2I Attn.' are transposed convolution and token-to-image attention, respectively. `LR Mask' and `HR Mask' denote the predicted low- and high-resolution mask logits, respectively. }
    \label{fig: s-decoder structure}
    \vspace{-2mm}
\end{figure}

The structure of the S-Decoder is presented in Fig.~\ref{fig: s-decoder structure}. The modules for predicting low-resolution masks in Fig.~\ref{fig: s-decoder structure} inherit the parameters of SAM's mask decoder. Let $\bm{t_{s\_out}} \in \mathbb{R}^{1 \times 256}$ denote the inherited output token, which is the first slice of SAM's output tokens. For ease of description, we omit another token which is used for IoU prediction here. The concatenated output and prompt tokens $[\bm{t_{s\_out}};\bm{t_{prompt}}] \in \mathbb{R}^{(1 + N) \times 256}$ are fed into S-Decoder. After the final token-to-image attention and transposed convolution, we obtain the updated output token $\bm{\hat{t}_{s\_out}} \in \mathbb{R}^{1 \times 256}$ and up-sampled mask features $\bm{F}\in\mathbb{R}^{256 \times 256 \times 32}$, where $256 \times 256$ is the default mask feature size in SAM, $32$ is the feature dimension. Then, $\bm{\hat{t}_{s\_out}}$ is passed to an MLP that outputs a vector matching the dimension of $\bm{F}$. The mask logits $\bm{M}$ for pixel-level text is predicted by a spatially point-wise product between $\bm{F}$ and the MLP's output. During inference, $\bm{M}$ will be interpolated into the input image size and further processed with a threshold.

%---------------------------------
%---------------------------------

For high-resolution mask logits, we reuse the updated output token $\bm{\hat{t}_{s\_out}}$ and mask features $\bm{F}$, while introducing a few lightweight modules. As illustrated in Fig.~\ref{fig: s-decoder structure}, we further up-sample $\bm{F}$ with another two transposed convolutional layers (kernel size $2 \times 2$, stride 2) and use four subsequent convolutional layers (kernel size $3 \times 3$, stride 1, padding 1) for refining the feature. After this step, we achieve the high-resolution mask feature $\bm{F_{hr}}$ with $1024\times 1024$ spatial size and $16$ dimensions. We initialize a three-layer MLP and apply it on $\bm{\hat{t}_{s\_out}}$ to produce a new vector. Similar to acquiring $\bm{M}$, S-Decoder predicts the final high-resolution mask logits $\bm{M_{hr}}$ of size $1024\times 1024$ based on $\bm{F_{hr}}$ and the updated vector.

In contrast to image super-resolution~\cite{chen2023activating}, where the input RGB image resolution is increased at the encoder stage, our approach focuses on enhancing the resolution specifically at the mask feature level within the decoder. This strategy reduces the computational overhead typically incurred by processing high-resolution images in the encoder.
Using SAM-TS-L as a case study, our method introduces high-resolution mask features at a minimal cost of 0.1 FPS (reducing from 2.9 FPS to 2.8 FPS on a single V100 GPU) and 0.79\% FLOPs (increasing from 1426.6G to 1437.8G), resulting in a 14.24 fgIOU improvement on HierText (see Tab.~\ref{tab: effect of HR}). Without high-resolution mask features, the ViT encoder would require processing $4096 \times 4096$ resolution input images to achieve the same segmentation mask resolution, leading to unaffordable out-of-memory issues.

\begin{figure}[!t]
    \centering
    \includegraphics[width=1\linewidth]{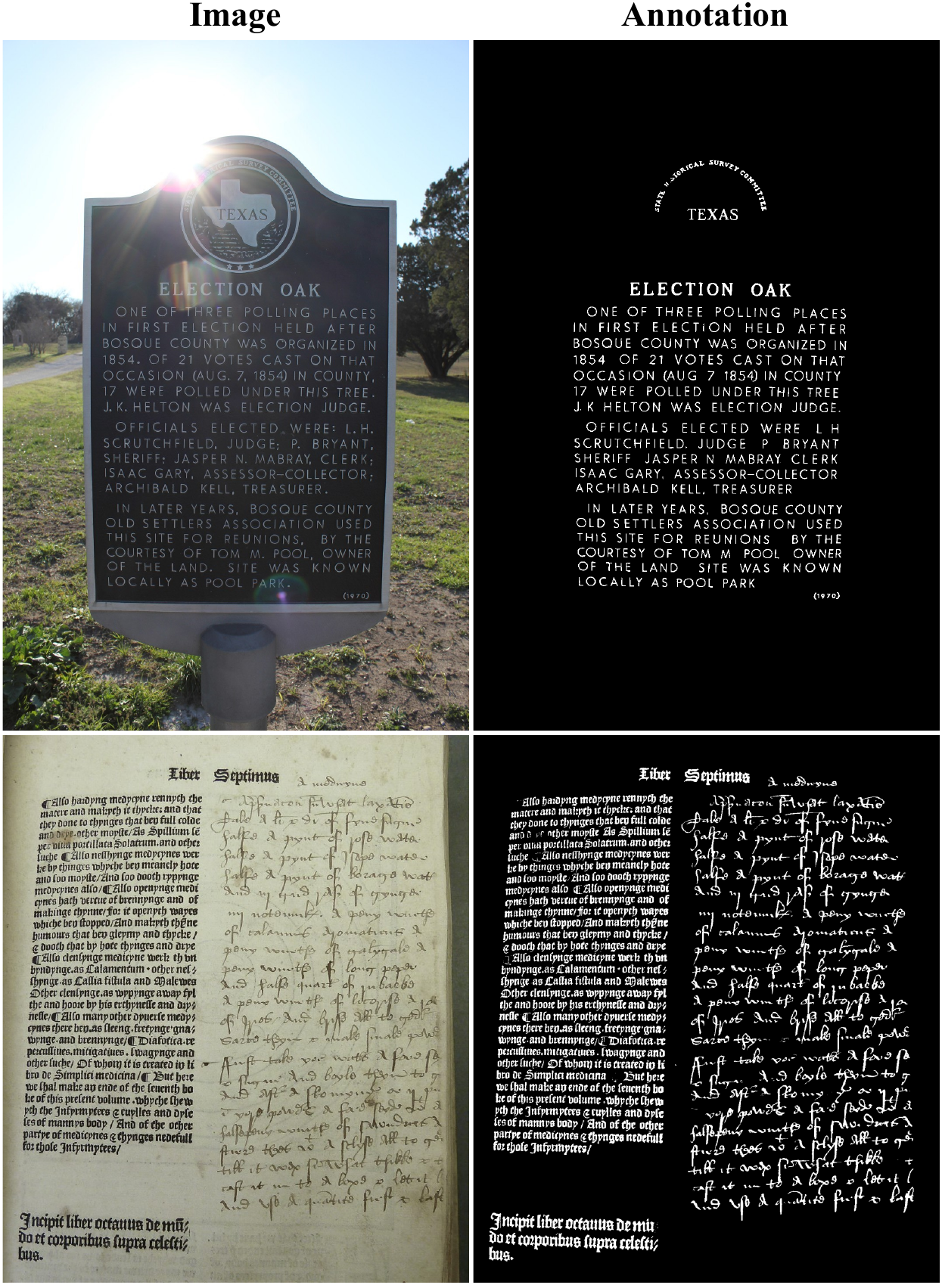}
    \caption{\textbf{Annotation samples in HierText generated by SAM-TS automatically.} Best view on screen with zooming in.}
    \label{fig: label_example}
    \vspace{-3mm}
\end{figure}

\noindent\textbf{Semi-automatic Annotation of Pixel-level Text Masks.}
We leverage the obtained SAM-TS model to generate mask labels for HierText in a semi-automatic way. Initially, we select some samples in document, poster, and handwritten scenarios with dense and fine-grained texts. We annotate them in Adobe Photoshop with tools such as contrast adjustment, reverse, brush, and binarization. After annotating 418 images in the HierText training set, 
we combine these images and the training images of Total-Text to train a SAM-TS model with the ViT-L backbone, \ie, SAM-TS-L.
During training, we use augmentations including color jittering, random blur, random rotation, and large-scale jittering~\cite{ghiasi2021simple}. We use this model to label the remaining training images of HierText. During inference, we leverage a sliding window strategy to improve segmentation quality. We set the window size to 512 $\times$ 512 with a stride of 384. False positive and false negative regions are modified manually. Specifically, we retrain the model and update the segmentation masks for the remaining images when the ratio between annotated images and total HierText training images reaches 1/8, 1/4, and 1/2. We only need to remove a few false positive segments as the model evolves. Finally, we use the latest model to label all the remaining images in the HierText training set. We also use SAM-TS-L to label the validation and testing images before manual inspection and modification. Fig.~\ref{fig: label_example} shows some annotation samples generated by SAM-TS-L.

%---------------------------------
%---------------------------------

\subsection{Word, Text-line, and Paragraph Segmentation}
\label{subsubsec: wtp segmentation}
Different from the S-Decoder which takes implicit prompts, the H-Decoder receives point prompts from SAM's prompt encoder. We obtain word, text-line, and paragraph masks in a single H-Decoder that is customized from SAM's mask decoder. Supposing we have $K$ foreground points on text, the prompt encoder processes them into prompt tokens $\bm{t_{point}} \in \mathbb{R}^{K \times N_p \times 256}$, where $N_p$ is a newly introduced feature dimension for prompt tokens, following the default prompt encoding process as in SAM. H-decoder takes the broadcasted output tokens $\bm{t_{h\_out}} \in \mathbb{R}^{K \times N_{out} \times 256}$ and point prompt tokens by concatenation ($[\bm{t_{h\_out}};\bm{t_{point}}] \in \mathbb{R}^{K \times (N_{out} + N_p) \times 256}$) as input. $N_{out}$ is the same output token number as in SAM's mask decoder. After the final token-to-image attention, we slice the last three output tokens and get $\bm{\hat{t}_{h\_out}} \in \mathbb{R}^{K \times 3 \times 256}$. In the original SAM's mask decoder, the last three output tokens are used for multi-mask output, without specific task assignment. In H-Decoder, for each point prompt, we employ the three output tokens to take charge of the word, text-line, and paragraph segmentation in order. Applying point-wise product between output tokens and mask features, we get the mask logits $\bm{M_{w,l,p}}$ in the shape of ${K \times 3 \times 256 \times 256}$, containing the corresponding word, text-line, and paragraph masks where the points are located in. 
Note that for each point prompt, the word mask contains not only the selected word but also other words in the same text line.
This is designed to ensure word recall during automatic mask generation for scenarios with dense text. UD~\cite{long2022towards} also adopts a similar design to ensure detection recall in its framework. In addition, to better separate adjacent word instances, the word output token is guided to segment word kernel regions. During inference, we follow the post-processing method in DB~\cite{liao2020real} to get complete word regions. In contrast, we do not conduct processing on text-line and paragraph labels.

Moreover, to improve the quality of word segmentation, we adopt a similar approach as in S-Decoder, \ie, generating mask features in higher resolution. Concretely, to save GPU memory usage, we use a convolutional layer with $1 \times 1$ kernel size to decrease the dimension of mask features from $32$ to $16$ at first. Then, we directly interpolate the mask features to have a spatial size of $384 \times 384$. We also use four convolutional layers with $3 \times 3$ kernel size to refine the mask features. A three-layer MLP is used to generate dynamic weights from the word-level output token. Finally, using the dynamic weights and refined mask features, we can get the word mask logits in higher resolution.

\subsection{Layout Analysis}
\label{subsubsec: layout}
Layout analysis in this work aims at clustering words into different blocks, where texts are visually and semantically coherent in each block. Intuitively, layout analysis requires mining relationships between text objects. 
Since the outputs from the H-Decoder inherently present a hierarchical relationship, layout analysis can be accomplished in passing, without the need of a specially designed branch. Specifically, in the predictions of the H-Decoder, each text-line mask is accompanied by a word mask (containing words in this text-line) and a paragraph mask. If the corresponding paragraph masks of different text-lines have high IoU, we regard these text-lines and their corresponding words in the same paragraph, thereby achieving layout analysis. To be specific, we calculate the IoU matrix ($\in [0, 1]^{K \times K}$) using paragraph masks in the shape of $K \times 256 \times 256$. Then, we group each pair of objects if their IoU metric exceeds $0.5$. A union-find algorithm used in~\cite{long2022towards} is adopted to merge these connected nodes into clusters.

\subsection{Training of Hi-SAM}
\label{subsubsec: training}
During training, the learnable modules in Hi-SAM contain 1) the adapters in the ViT encoder, 2) the self-prompting module, 3) the S-Decoder, and 4) the H-Decoder. We optimize them in an end-to-end manner. In terms of the S-Decoder, we supervise both the low-resolution and high-resolution mask predictions. The loss function $\mathcal{L}_{lr}$ for low-resolution part is a linear combination of Focal loss~\cite{lin2017focal}, Dice loss~\cite{milletari2016v}, and IoU mean-square-error (MSE) loss in a ratio of 20:1:1, following~\cite{kirillov2023segment}. Similarly, the loss function $\mathcal{L}_{hr}$ for the high-resolution part is also a linear combination of Focal loss, Dice loss, and MSE loss in the same ratio. For pixel-level text segmentation, the overall loss: $\mathcal{L}_{text} = \mathcal{L}_{lr} + \mathcal{L}_{hr}$.

In terms of the H-Decoder, we employ a training strategy contrapuntally. For each image which may contain up to hundreds of text-lines, we assume that some text-lines are similar to each other in appearance. We only use some samples for training the H-Decoder. Specifically, we randomly sample at most 10 text-lines in each image. Then, we calculate the intersection of text-line masks and the pixel-level text masks to determine the foreground points in each text-line. For each text-line, we randomly sample 2 foreground points. At last, we can get at most 20 foreground points along with the ground-truth word, text-line, and paragraph masks. 
We feed foreground points into the prompt encoder to generate prompt tokens for the H-Decoder. If there is no foreground point, the H-Decoder is guided to segment blank masks. Overall, for word segmentation, the loss function is defined as $\mathcal{L}_{word} = \mathcal{L}_{w\_lr} + \mathcal{L}_{w\_hr}$, where $\mathcal{L}_{w\_lr}$ and $\mathcal{L}_{w\_hr}$ are used to supervise the low-resolution and high-resolution word mask predictions. Each of them contains binary-cross-entropy (BCE) loss and Dice loss in a ratio of 1:1, following~\cite{sam_hq}. For text-line, the loss $\mathcal{L}_{line}$ is a linear combination of BCE loss, Dice loss, and IoU MSE loss in a ratio of 1:1:1. We use the same setting for paragraph loss $\mathcal{L}_{para}$. The predicted scores from IoU heads can be used to conduct mask non-maximum-suppression (NMS) during auto-mask generation. The final loss function $\mathcal{L}$ is formulated as:
\begin{equation}
    \mathcal{L} = \mathcal{L}_{text} + \mathcal{L}_{word} + \mathcal{L}_{line} + 0.5 \times \mathcal{L}_{para}.
\end{equation}

\subsection{Inference of Hi-SAM}
\label{subsubsec: inference}
\noindent\textbf{Automatic Mask Generation.} 
For each input image, SAM-TS first generates pixel-level text mask, where the pixels of text are considered as foreground. Then, we randomly sample $P$ foreground points, which are embedded into prompt tokens via the prompt encoder. H-Decoder uses them to predict dense word, text-line, and paragraph masks. We regard text-line as a pivot. We filter out the text-line masks whose predicted IoU scores are lower than $0.5$. Their corresponding word and paragraph masks are also discarded. Next, we conduct Matrix NMS~\cite{wang2020solov2} on GPU using the remaining text-line masks and their IoU scores. The threshold for Matrix NMS is set to $0.5$ by default. After NMS, we gather the final text-line masks, and their corresponding word and paragraph masks. Finally, we accomplish layout analysis based on the paragraph masks as described in Sec.~\ref{subsubsec: layout}. An additional NMS at the paragraph level can be applied if non-redundant paragraph masks are required. All the predicted masks can be interpolated into the same resolution as the input image. In this way, we achieve hierarchical text segmentation in a unified framework, and realize layout analysis in passing.

\noindent\textbf{Promptable Segmentation.} In this mode, the inference pipeline only involves the image encoder, prompt encoder, and H-decoder. The image encoder is used to extract embedding at first. For the points clicked by a user, the prompt encoder embeds them into prompt tokens and the H-Decoder generates word, text-line, and paragraph masks for each point. Since each word mask contains the words in one text-line as mentioned in Sec.~\ref{subsubsec: wtp segmentation}, we compare the proximity between the clicked point and word instances to determine the selected word.

\section{Experiments}
\label{sec:experiments}
\subsection{Datasets and Evaluation Protocols}
\label{subsec: datasets}
\noindent\textbf{Total-Text}~\cite{ch2017total} includes 1,255 training images and 300 testing images. It contains scene texts with arbitrary shapes. Total-Text provides binary mask labels of pixel-level text and polygon annotations of word instances.

\noindent\textbf{TextSeg}~\cite{xu2021rethinking} contains 4,024 images, which are split into training, validation, and testing sets with 2,646, 340, and 1,038 images, respectively. It focuses on both scene texts and designed texts for pixel-level text segmentation.

\noindent\textbf{COCO\_TS}~\cite{bonechi2019coco_ts} has 14,690 images from COCO-Text~\cite{veit2016coco}.
The COCO\_TS pixel-level text masks are derived automatically from bounding box annotations in COCO-Text using a weakly supervised segmentation model trained on extensive synthetic data. Due to the limited labeling quality, we exclusively employ COCO\_TS as a reference to evaluate the automatic labeling capability of our SAM-TS model.

\noindent\textbf{HierText}~\cite{long2022towards} consists of 8,281 training images, 1,724 validation images, and 1,634 testing images, with dense and small texts in scene, document, \etc. It contains hierarchical word, text-line, and paragraph location annotations. Word locations are annotated with polygons while text-line locations are annotated with quadrilateral boxes. Paragraph locations are represented by coarse polygons. HierText features scene, designed, printed, and handwritten texts. We contribute the annotations for pixel-level text segmentation in this paper.

We use the IOU and F-score metrics of foreground pixels for pixel-level text segmentation~\cite{xu2021rethinking}. We follow HierText~\cite{long2022towards} to evaluate word, text-line, paragraph (layout) tasks with Panoptic Quality (PQ)~\cite{kirillov2019panoptic}, F-score, Precision (P), Recall (R), and Tightness (T). PQ and F-score are primary metrics.

%---------------------------------
\begin{figure}[!t]
    \centering
    \includegraphics[width=\linewidth]{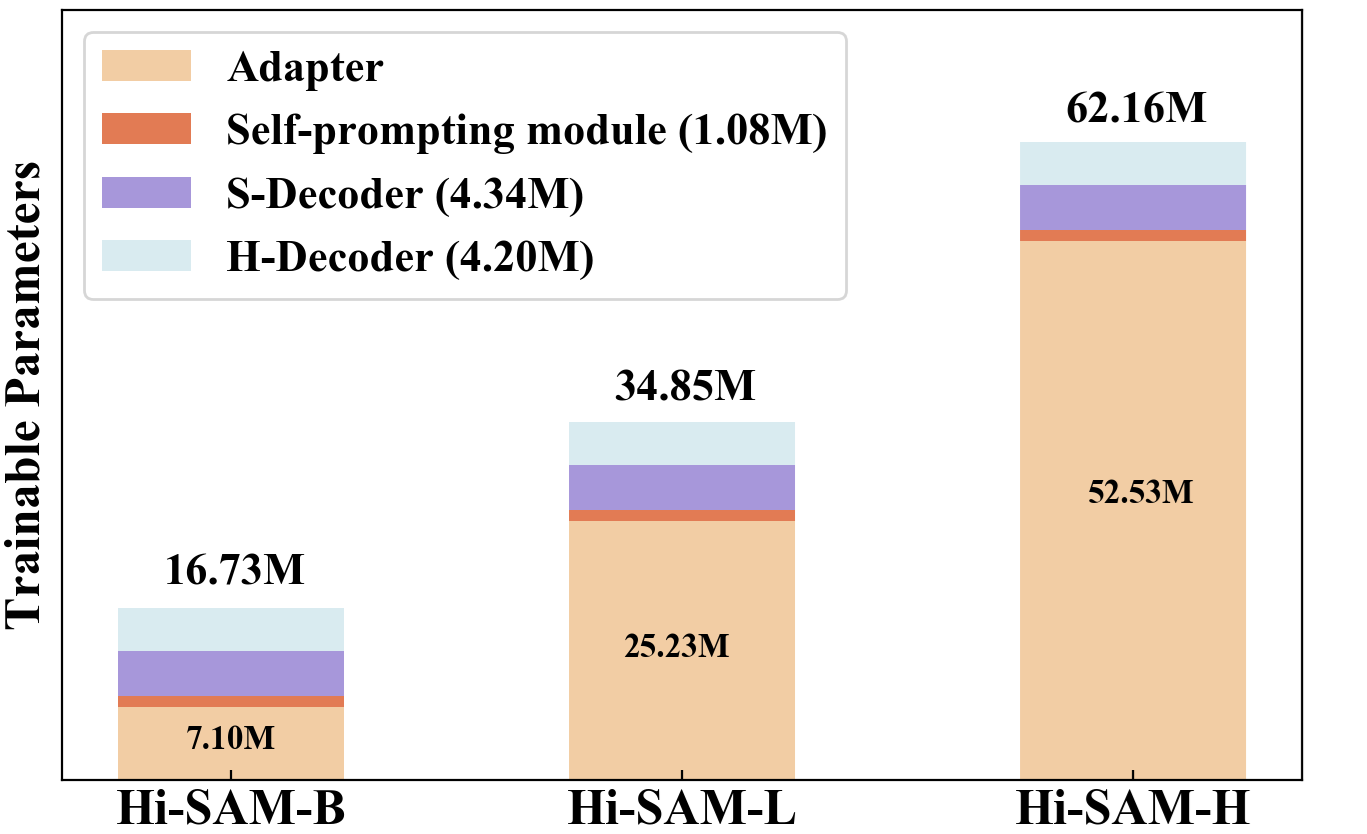}
    \caption{\textbf{Trainable parameter statistics of Hi-SAM with ViT-B, ViT-L, ViT-H backbones, respectively.}}
    \label{fig: params_hisam}
\end{figure}
%---------------------------------

\subsection{Implementation Details}
\noindent\textbf{Training and Inference on Benchmarks.}
Hi-SAM takes an input image size of $1024 \times 1024$, which is the same as SAM~\cite{kirillov2023segment}. In our default setting, the self-prompting module generates 12 prompt tokens for the S-Decoder. We use AdamW~\cite{loshchilov2017decoupled} ($\beta_1 = 0.9, \beta_2 = 0.999, weight\_decay = 0.05$) as the optimizer. The learning rate is set to $1e^{-4}$. The batch size is 8. We augment the input image with color jittering, random rotation, and large-scale jittering with a scale range of [0.5, 2.0].
For TS, we compare SAM-TS with existing representative TS methods. During training, for each dataset, we only use its training set following previous methods. On Total-Text and TextSeg, we train SAM-TS for 70 epochs without a learning rate drop. On HierText, we train SAM-TS for 80 epochs with the learning rate divided by 10 at 70 epochs. During inference, the input image size is also $1024 \times 1024$.
For hierarchical text segmentation, we only use the training set of HierText for training. We train Hi-SAM for 150 epochs with the learning rate divided by 10 at 130 epochs. During training, we randomly sample at most 10 text-lines in each image and 2 points on each text-line. The sampling point number $P$ in Sec.~\ref{subsubsec: inference} is set to 1,500. In practice, we split these points into batches, with 100 points per batch in the default setting. We train the models on 8 NVIDIA Tesla V100 (32GB) GPUs.

\noindent\textbf{Trainable Parameter Statistics of Hi-SAM.}
We provide the trainable parameter statistics of Hi-SAM with different backbones in Fig.~\ref{fig: params_hisam}. Hi-SAM-B, Hi-SAM-L, and Hi-SAM-H employ ViT-B, ViT-L, and ViT-H backbones, respectively. 
Specifically, in Hi-SAM-B, there are 16.73M trainable parameters in total, with 7.10M in the adapter. In the three Hi-SAM variants, the self-prompting module only has 1.08M trainable parameters, while the S-Decoder and H-Decoder account for 4.34M and 4.20M trainable parameters, respectively.
Note that SAM-TS-B for TS does not need an H-Decoder and has 12.53M trainable parameters in total, while SAM-TS-L and SAM-TS-H have 30.65M and 57.95M trainable parameters, respectively.

\begin{table}[t!]
    \centering
    \setlength{\tabcolsep}{5pt}
    \caption{
    \textbf{Influence of the adapter in backbone and the Transformer decoder layer in self-prompting on Total-Text.}
    }
    \label{tab: adapter&tdeclayer}
    \resizebox{0.8\linewidth}{!}{
    \begin{tabular}{c|cc|cc}
         \toprule[1pt]
         \rowcolor{gray!20} \textbf{Backbone} &\textbf{Adapter} &\textbf{TDecLayer} &\textbf{fgIOU} &\textbf{F-score} \\
         \midrule[1pt]
         \multirow{3}{*}{ViT-B} & & &68.80 &78.31 \\
         &\checkmark & &79.73 &86.25 \\
         &\checkmark &\checkmark &\textbf{80.93} &\textbf{86.25} \\
         \midrule[0.5pt]
         \multirow{3}{*}{ViT-L} & & &69.17 &79.86 \\
         &\checkmark & &84.02 &88.60 \\
         &\checkmark &\checkmark &\textbf{84.59} &\textbf{88.69} \\
         \bottomrule[1pt]
    \end{tabular}}
\end{table}

\begin{table}[t!]
    \centering
    \setlength{\tabcolsep}{10pt}
    \caption{
    \textbf{Influence of prompt token number on Total-Text.}
    }
    \label{tab: token num}
    \resizebox{0.8\linewidth}{!}{
    \begin{tabular}{c|c|cc}
         \toprule[1pt]
         \rowcolor{gray!20} \textbf{Backbone} &\textbf{Tokens} &\textbf{fgIOU} &\textbf{F-score} \\
         \midrule[1pt]
         \multirow{3}{*}{ViT-B} &8 &80.02 &85.98 \\
         &12 &80.93 &\textbf{86.25} \\
         &16 &\textbf{81.04} &86.22 \\
         \midrule[0.5pt]
         \multirow{3}{*}{ViT-L} &8 &83.92 &88.64 \\
         &12 &\textbf{84.59} &\textbf{88.69} \\
         &16 &83.94 &88.44 \\
         \bottomrule[1pt]
    \end{tabular}}
\end{table}

\subsection{Ablation Studies}
\label{subsec: ablation}

\subsubsection{Pixel-level Text Segmentation}
\label{subsubsec: ablation_TS_seg}

\noindent\textbf{Influence on the Adapter and Transformer Decoder Layer in the Self-prompting Module.}
As shown in Tab.~\ref{tab: adapter&tdeclayer}.
We find that directly applying the frozen image encoder of SAM failed to yield satisfactory segmentation results, suggesting that the original SAM lacks awareness of fine text details. By introducing a few trainable parameters in the ViT blocks, both ViT-B and ViT-L models achieve significantly better performance than the baseline, validating the importance of adapting SAM to the text image domain. Moreover, using a single Transformer decoder layer to refine the tokens in the self-prompting module can further bring improvements.

\noindent\textbf{Influence of Prompt Token Number.} We investigate the influence of prompt token number on ViT-B and ViT-L models in Tab.~\ref{tab: token num}. For ViT-B, using 12 prompt tokens achieves the best F-score and comparable fgIOU as using 16 prompt tokens, while using 12 prompt tokens for ViT-L achieves the best performance. 
For a better trade-off between complexity and performance, we adopt 12 prompt tokens by default for pixel-level text segmentation.

\begin{table}[t!]
    \centering
    \setlength{\tabcolsep}{8pt}
    \caption{
    \textbf{Comparison of different token representations on Total-Text.}
    }
    \label{tab: comparison of token}
    \resizebox{\linewidth}{!}{
    \begin{tabular}{c|c|cc}
         \toprule[1pt]
         \rowcolor{gray!20} \textbf{Backbone} &\textbf{Method} &\textbf{fgIOU} &\textbf{F-score} \\
         \midrule[1pt]
         \multirow{2}{*}{ViT-B} &vanilla embedding~\cite{carion2020end} &80.78 &85.95 \\
         &spatial attention~\cite{ryoo2021tokenlearner} &\textbf{80.93} &\textbf{86.25} \\
         \midrule[0.5pt]
         \multirow{2}{*}{ViT-L} &vanilla embedding~\cite{carion2020end} &83.92 &88.32 \\
         &spatial attention~\cite{ryoo2021tokenlearner} &\textbf{84.59} &\textbf{88.69} \\
         \bottomrule[1pt]
    \end{tabular}}
\end{table}

\begin{table}[t!]
    \centering
    \setlength{\tabcolsep}{3pt}
    \caption{
    \textbf{Effectiveness of introducing high-resolution mask features in S-Decoder.} `HR' denotes whether equipped with high-resolution mask features. `fgIOU-LR' means the fgIOU evaluated with the outputs from low-resolution mask features while `fgIOU-HR' is achieved from high-resolution features.
    }
    \label{tab: effect of HR}
    \resizebox{\linewidth}{!}{
    \begin{tabular}{c|c|cc|cc}
         \toprule[1pt]
         \rowcolor{gray!20}& &\multicolumn{2}{c|}{\textbf{Total-Text}} &\multicolumn{2}{c}{\textbf{HierText}} \\
         \cline{3-4} \cline{5-6} 
         \rowcolor{gray!20} \multirow{-2}{*}{\textbf{Backbone}} &\multirow{-2}{*}{\textbf{HR}} &\textbf{fgIOU-LR} &\textbf{fgIOU-HR} &\textbf{fgIOU-LR} &\textbf{fgIOU-HR} \\
         \midrule[1pt]
         \multirow{2}{*}{ViT-B} & &79.70 &$-$ &61.58 &$-$ \\
         &\checkmark &79.91 &80.93~(\textcolor[rgb]{0,0.8,0}{+1.23}) &61.63 &73.39~(\textcolor[rgb]{0,0.8,0}{+11.81}) \\
         \midrule[0.5pt]
         \multirow{2}{*}{ViT-L} & &82.28 &$-$ &64.13 &$-$ \\
         &\checkmark &83.40 &84.59~(\textcolor[rgb]{0,0.8,0}{+2.31}) &63.75 &78.37~(\textcolor[rgb]{0,0.8,0}{+14.24}) \\
         \midrule[1pt]
    \end{tabular}}
\end{table}

%---------------------------------
\begin{figure}[!t]
    \centering
    \includegraphics[width=1\linewidth]{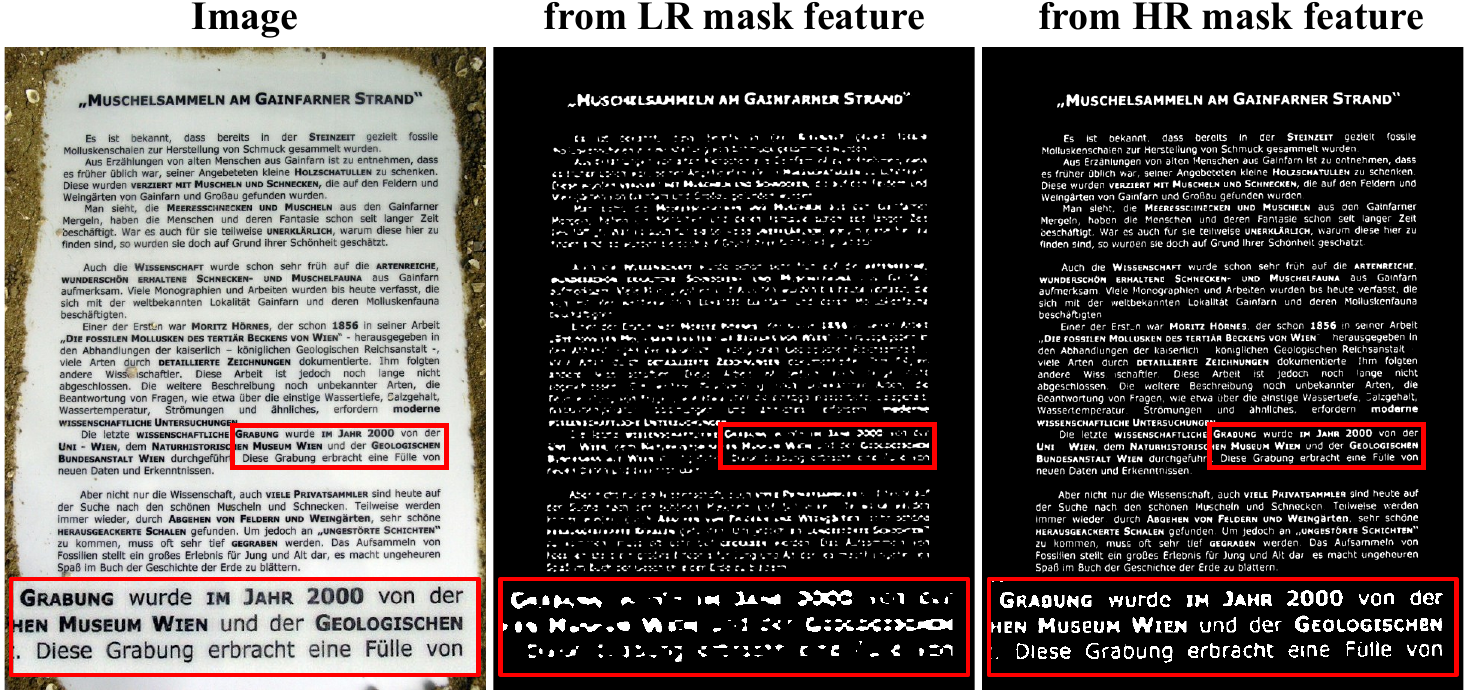}
    \caption{\textbf{Qualitative comparison between the predictions from low-resolution (LR) mask feature and high-resolution (HR) mask feature.}}
    \label{fig: comp_LR_HR}
\end{figure}
%---------------------------------

\noindent\textbf{Comparison of Different Token Representation Methods.}
In the self-prompting module, we adopt a lightweight tokenizer~\cite{ryoo2021tokenlearner} based on spatial attention to extract initial prompt tokens from image embeddings. As presented in Tab.~\ref{tab: comparison of token}, we compare it with vanilla embedding, which uses initialized embeddings as prompt tokens, similar to the object queries in DETR~\cite{carion2020end}.
After training, the prompt tokens derived from vanilla embedding are fixed while the ones from spatial attention are related to image embeddings. We observe that the latter method performs better, especially when image embeddings become more representative. Specifically, compared with the vanilla embedding, the latter gets 0.15\% and 0.30\% improvements in terms of fgIOU and F-score with ViT-B, while achieving 0.67\% and 0.37\% improvements on fgIOU and F-score with ViT-L.

\noindent\textbf{Effectiveness of High-resolution Mask Features in S-Decoder.}
In Tab.~\ref{tab: effect of HR}, we validate the effectiveness of using high-resolution (HR) mask features for pixel-level text segmentation. We observe that introducing HR mask features significantly boosts the segmentation quality. With ViT-L, the fgIOU performance is improved by 2.31\% on Total-Text. Moreover, on HierText which requires segmenting dense texts in scenes, documents, and handwritten materials, leveraging HR mask features delivers remarkable 14.24\% fgIOU improvements.
Some qualitative results are provided in Fig.~\ref{fig: comp_LR_HR}. The result of using the LR mask feature almost loses all stroke details. In comparison, the prediction from the HR mask reserves most details.

\noindent\textbf{Influence of the tuning method and pretrained weights for initialization.}
In Tab.~\ref{tab: influence of SAM on text seg}, we demonstrate that adapter-tuning significantly enhances model adaptation to TS on smaller datasets, particularly with the larger ViT-L backbone. Additionally, when using MAE pretrained ViT weights instead of SAM's weights (for both the image encoder and mask decoder) for initialization, the fgIOU of SAM-TS-B and SAM-TS-L decreases by 7.21\% and 5.13\%, respectively.
The results indicate that SAM effectively learns general segmentation knowledge through large-scale training, which can be successfully transferred to pixel-level text segmentation to improve performance.

\subsubsection{Hierarchical Text Segmentation and Layout Analysis}
\label{subsubsec: ablation_hi_seg}
In this part, we focus on the ablation studies about the H-Decoder, training, and inference setting of Hi-SAM. We conduct experiments on the validation set of HierText using ViT-L as the backbone.

\begin{table}[t!]
    \centering
    \setlength{\tabcolsep}{3pt}
    \caption{
    \textbf{Influence of the tuning method and pretrained weights for initialization on Total-Text.} `model-tuning' means all parameters are fully fine-tuned, without the adapter plugged in. `$-$' means the whole model is trained from scratch.
    }
    \label{tab: influence of SAM on text seg}
    \resizebox{\linewidth}{!}{
    \begin{tabular}{c|cc|cc}
         \toprule[1pt]
         \rowcolor{gray!20} \textbf{Backbone} &\textbf{Method} &\textbf{Pretrained Weights} &\textbf{fgIOU} &\textbf{F-score} \\
         \midrule[1pt]
         \multirow{4}{*}{ViT-B} &adapter-tuning &SAM &\textbf{80.93} &\textbf{86.25} \\
         &model-tuning &SAM &80.27 &85.98 \\
         &model-tuning &MAE pretrained ViT &73.06 &83.24 \\
         &model-tuning &$-$ &41.03 &57.70 \\
         \midrule[0.5pt]
         \multirow{4}{*}{ViT-L} &adapter-tuning &SAM &\textbf{84.59} &\textbf{88.69} \\
         &model-tuning &SAM &80.52 &86.68 \\
         &model-tuning &MAE pretrained ViT &75.39 &84.55 \\
         &model-tuning &$-$ &41.86 &58.85 \\
         \bottomrule[1pt]
    \end{tabular}}
\end{table}

\begin{table}[t!]
    \centering
    \caption{
    \textbf{Effectiveness of introducing high-resolution mask features in the H-Decoder.} `Feature Size' denotes the word mask feature size. The mask feature size for text-line and paragraph segmentation are both set to $256 \times 256$.
    }
    \label{tab: effect of HR in h-decoder}
    \setlength{\tabcolsep}{4pt}
    \resizebox{\linewidth}{!}{
    \begin{tabular}{c|cc|cc|cc}
         \toprule[1pt]
         \rowcolor{gray!20} &\multicolumn{2}{c|}{\textbf{Word}} &\multicolumn{2}{c|}{\textbf{Text-line}} &\multicolumn{2}{c}{\textbf{Layout Analysis}} \\
         \cline{2-3} \cline{4-5} \cline{6-7} 
         \rowcolor{gray!20} \multirow{-2}{*}{\textbf{Feature Size}} &\textbf{PQ} &\textbf{F-score} &\textbf{PQ} &\textbf{F-score} &\textbf{PQ} &\textbf{F-score} \\
         \midrule[1pt]
         $256 \times 256$ &60.15 &78.83 &64.72 &83.71 &58.55 &75.80 \\
         $384 \times 384$ &\textbf{62.60} &\textbf{80.99} &\textbf{65.62} &\textbf{84.04} &\textbf{58.83} &\textbf{75.84} \\
         % \midrule[0.5pt]
         $\Delta$ &\textcolor[rgb]{0,0.8,0}{+2.45} &\textcolor[rgb]{0,0.8,0}{+2.16} &\textcolor[rgb]{0,0.8,0}{+0.90} &\textcolor[rgb]{0,0.8,0}{+0.33} &\textcolor[rgb]{0,0.8,0}{+0.28} &\textcolor[rgb]{0,0.8,0}{+0.04} \\
         \midrule[1pt]
    \end{tabular}}
\end{table}

\noindent\textbf{Effectiveness of High-resolution Mask Features in H-Decoder.}
As described in Sec.~\ref{subsubsec: wtp segmentation}, we also introduce the mask feature with higher resolution to improve the quality of word segmentation in the H-Decoder. The performance comparison between different feature resolutions is listed in Tab.~\ref{tab: effect of HR in h-decoder}. As can be seen, utilizing the high-resolution mask feature in the H-Decoder leads to better capability in distinguishing adjacent word instances on book pages and document materials, bringing ideal improvement on word segmentation, \ie, 2.45\% PQ and 2.16\% F-score. Note that the metrics for text-line and layout analysis are also enhanced because these tasks regard word instance as the basic unit.

\noindent\textbf{Impact of Different Sample Strategies.}
During training, we randomly sample 10 text-lines in each image and 2 points on each text-line. We also examine the influence of increasing the point number on each text-line or adding text-line samples. The comparison results and corresponding training costs are shown in Tab.~\ref{tab: h-decoder training strategy}.
As we can see while keeping the text-lines at 10 and increasing the point amount to 5, the PQ and F-score improved by 0.54\% and 0.73\% at the word level, 0.33\% and 0.21\% at the text-line level, and 0.63\% and 0.48\% for layout analysis. In comparison, increasing text-line samples achieves more gains in layout analysis. However, increasing training samples requires more training resources. For example, only increasing the point amount on each text-line from 2 to 5 results in about 1.2$\times$ training time and 6GB more memory footprint on each GPU. Sampling more text-lines also leads to more data processing time. Therefore, we use 10 text-lines per image and 2 points per line as the default setting.

\begin{table}[t!]
    \centering
    \caption{
    \textbf{Comparison of different training strategies.} `10Ls $\times$ 2Pts' means sampling 10 text-lines in each image and 2 points on each text-line, resulting in at most 20 foreground points for prompt encoder and H-Decoder. `Time' denotes the minutes per epoch. `Mem.' denotes the peak memory footprint per GPU.
    }
    \label{tab: h-decoder training strategy}
    \setlength{\tabcolsep}{2pt}
    \resizebox{\linewidth}{!}{
    \begin{tabular}{c|cc|cc|cc|c}
         \toprule[1pt]
         \rowcolor{gray!20} &\multicolumn{2}{c|}{\textbf{Word}} &\multicolumn{2}{c|}{\textbf{Text-line}} &\multicolumn{2}{c|}{\textbf{Layout Analysis}} & \\
         \cline{2-3} \cline{4-5} \cline{6-7} 
         \rowcolor{gray!20} \multirow{-2}{*}{\textbf{Strategy}} &\textbf{PQ} &\textbf{F-score} &\textbf{PQ} &\textbf{F-score} &\textbf{PQ} &\textbf{F-score} &\multirow{-2}{*}{\textbf{Time (Mem.)}}\\
         \midrule[1pt]
         10Ls $\times$ 2Pts &62.60 &80.99 &65.62 &84.04 &58.83 &75.84 &20.5min (16GB) \\
         10Ls $\times$ 5Pts &\textbf{63.14} &\textbf{81.72} &\textbf{65.95} &\textbf{84.25} &59.46 &76.32 &24.0min (22GB) \\
         25Ls $\times$ 2Pts &63.02 &81.57 &65.81 &84.12 &\textbf{60.03} &\textbf{77.10} &32.0min (22GB) \\
         \midrule[1pt]
    \end{tabular}}
\end{table}

\begin{table}[t!]
    \centering
    \caption{
    \textbf{Comparison of different point sampling number $P$ during automatic mask generation.}
    }
    \label{tab: h-decoder inference points}
    \setlength{\tabcolsep}{4pt}
    \resizebox{\linewidth}{!}{
    \begin{tabular}{c|cc|cc|cc}
         \toprule[1pt]
         \rowcolor{gray!20} &\multicolumn{2}{c|}{\textbf{Word}} &\multicolumn{2}{c|}{\textbf{Text-line}} &\multicolumn{2}{c}{\textbf{Layout Analysis}} \\
         \cline{2-3} \cline{4-5} \cline{6-7} 
         \rowcolor{gray!20} \multirow{-2}{*}{\textbf{Points}} &\textbf{PQ} &\textbf{F-score} &\textbf{PQ} &\textbf{F-score} &\textbf{PQ} &\textbf{F-score} \\
         \midrule[1pt]
         500 &61.83 &79.82 &63.96 &81.63 &57.49 &74.01 \\
         1,000 &62.46 &80.76 &65.28 &83.52 &58.55 &75.44 \\
         1,500 &62.60 &80.99 &65.62 &84.04 &58.83 &75.84 \\
         2,000 &\textbf{62.64} &\textbf{81.07} &\textbf{65.74} &\textbf{84.24} &\textbf{58.92} &\textbf{75.97} \\
         \midrule[1pt]
    \end{tabular}}
\end{table}

\begin{table}[t!]
    \centering
    \caption{
    \textbf{Influence of the tuning method and pretrained weights for initialization on Hi-SAM.}
    }
    \label{tab: influence of SAM on Hi-SAM}
    \setlength{\tabcolsep}{1pt}
    \resizebox{\linewidth}{!}{
    \begin{tabular}{cc|cc|cc|cc}
         \toprule[1pt]
         \rowcolor{gray!20} & &\multicolumn{2}{c|}{\textbf{Word}} &\multicolumn{2}{c|}{\textbf{Text-line}} &\multicolumn{2}{c}{\textbf{Layout Analysis}} \\
         \cline{3-4} \cline{5-6} \cline{7-8} 
         \rowcolor{gray!20} \multirow{-2}{*}{\textbf{Method}} &\multirow{-2}{*}{\textbf{Pretrained Weights}} &\textbf{PQ} &\textbf{F-score} &\textbf{PQ} &\textbf{F-score} &\textbf{PQ} &\textbf{F-score} \\
         \midrule[1pt]
         adapter-tuning &SAM &\textbf{62.60} &\textbf{80.99} &\textbf{65.62} &\textbf{84.04} &\textbf{58.83} &\textbf{75.84} \\
         model-tuning &SAM &60.85 &78.99 &63.55 &81.71 &58.12 &75.03 \\
         model-tuning &MAE pretrained ViT &57.67 &75.42 &61.39 &79.95 &56.05 &73.21 \\
         model-tuning &$-$ &47.13 &63.57 &52.12 &70.21 &47.44 &63.83 \\
         \midrule[1pt]
    \end{tabular}}
    \vspace{-3mm}
\end{table}

\noindent\textbf{Influence of Point Sampling Number $P$.}
As mentioned in Sec.~\ref{subsubsec: inference}, we randomly sample $P$ foreground points as the inputs for the prompt encoder. We evaluate the influence of using different point sampling numbers in Tab.~\ref{tab: h-decoder inference points}. The performance of sampling 500 points is relatively low since it is insufficient to represent texts in HierText which has an average of 103.8 word instances per image. When the sampling points are doubled, the F-score improves by 0.94\%, 1.89\%, and 1.43\% on word, text-line, and layout analysis, respectively. The performance could be further improved by using more points but it saturates at 2,000 points. For a better trade-off between performance and inference overhead, we set the point sampling number $P$ to 1,500 by default.

\begin{table}[t!]
    \centering
    \setlength{\tabcolsep}{1pt}
    \caption{
    \textbf{Comparison results of pixel-level text segmentation on Total-Text.} The result of SegFormer is cited from TFT~\cite{yu2023scene}. The best and second best are marked with \textbf{bold} and \underline{underline}. `Tr.Params' denotes the trainable parameters. `To.Params' denotes the total parameters. They have the same meanings for other tables. Previous methods do not leverage the foundation model. SAM-TS leverages SAM in a parameter-efficient training manner, thus it owns more total parameters but fewer trainable parameters.
    }
    \label{tab: comp on totaltext}
    \resizebox{\linewidth}{!}{
    \begin{tabular}{l|cc|cc}
         \toprule[1pt]
         \rowcolor{gray!20} \textbf{Method} &\textbf{Tr.Params} &\textbf{To.Params} &\textbf{fgIOU} &\textbf{F-score} \\
         \midrule[1pt]
         DeepLabV3+~\cite{chen2018encoder} &59.3M &59.3M &74.44 &82.40 \\
         HRNetV2-W48~\cite{wang2020deep} &65.9M &65.9M &75.29 &82.50 \\
         HRNetV2-W48 + OCR~\cite{wang2020deep} &70.3M &70.3M &76.23 &83.20 \\
         TexRNet + DeeplabV3+~\cite{xu2021rethinking} &59.9M &59.9M &76.53 &84.40 \\
         TexRNet + HRNetV2-W48~\cite{xu2021rethinking} &67.1M &67.1M &78.47 &84.80 \\
         SegFormer~\cite{xie2021segformer} &$-$ &$-$ &73.31 &84.60 \\
         PGTSNet~\cite{xu2022bts} &$-$ &$-$ &79.10 &84.70 \\
         TFT~\cite{yu2023scene} &$-$ &$-$ &82.10 &\textbf{90.20} \\
         \midrule[0.5pt]
         SAM-TS-B &12.5M &102.2M &80.93 &86.25 \\
         SAM-TS-L &30.7M &338.9M &\underline{84.59} &88.69 \\
         SAM-TS-H &58.0M &695.0M &\textbf{84.86} &\underline{89.68} \\
         \bottomrule[1pt]
    \end{tabular}}
\end{table}

\begin{table}[t!]
    \centering
    \setlength{\tabcolsep}{1pt}
    \caption{
    \textbf{Comparison results of pixel-level text segmentation on TextSeg.} The result of SegFormer is cited from TFT~\cite{yu2023scene}.
    }
    \label{tab: comp on textseg}
    \resizebox{\linewidth}{!}{
    \begin{tabular}{l|cc|cc}
         \toprule[1pt]
         \rowcolor{gray!20} \textbf{Method} &\textbf{Tr.Params} &\textbf{To.Params} &\textbf{fgIOU} &\textbf{F-score} \\
         \midrule[1pt]
         DeepLabV3+~\cite{chen2018encoder} &59.3M &59.3M &84.07 &91.40 \\
         HRNetV2-W48~\cite{wang2020deep} &65.9M &65.9M &85.03 &91.40 \\
         HRNetV2-W48 + OCR~\cite{wang2020deep} &70.3M &70.3M &85.98 &91.80 \\
         TexRNet + DeeplabV3+~\cite{xu2021rethinking} &59.9M &59.9M &86.06 &92.10 \\
         TexRNet + HRNetV2-W48~\cite{xu2021rethinking} &67.1M &67.1M &86.84 &92.40 \\
         SegFormer~\cite{xie2021segformer} &$-$ &$-$ &84.59 &91.60 \\
         TFT~\cite{yu2023scene} &$-$ &$-$ &87.11 &93.10 \\
         \midrule[0.5pt]
         SAM-TS-B &12.5M &102.2M &87.15 &92.81 \\
         SAM-TS-L &30.7M &338.9M &\underline{88.77} &\underline{93.79} \\
         SAM-TS-H &58.0M &695.0M &\textbf{88.96} &\textbf{93.87} \\
         \bottomrule[1pt]
    \end{tabular}}
\end{table}

\begin{table}[t!]
    \centering
    \setlength{\tabcolsep}{1pt}
    \caption{
    \textbf{Comparison results of pixel-level text segmentation on HierText.} We train TextRNet~\cite{xu2021rethinking} with their official codes. `S: 1024': the shorter side of each image is resized to 1024 and the aspect ratio is kept. `\dag': applying sliding window strategy (window size: 512 $\times$ 512, stride: 384).
    }
    \label{tab: comp on hiertext}
    \resizebox{\linewidth}{!}{
    \begin{tabular}{l|cc|cc}
         \toprule[1pt]
         \rowcolor{gray!20} \textbf{Method} &\textbf{Tr.Params} &\textbf{To.Params} &\textbf{fgIOU} &\textbf{F-score}\\
         \midrule[1pt]
         TexRNet + HRNetV2-W48~\cite{xu2021rethinking} (S: 1024) &67.1M &67.1M &55.50 &65.64 \\
         TexRNet + HRNetV2-W48~\cite{xu2021rethinking} (S: 1536) &67.1M &67.1M &65.72 &75.19 \\
         TexRNet + HRNetV2-W48~\cite{xu2021rethinking} (S: 2048) &67.1M &67.1M &70.77 &80.32 \\
         % TexRNet + HRNetV2-W48~\cite{xu2021rethinking} (S: 4096) &67.1M &67.1M &72.25 &84.81 \\ %FPS 0.4
         \midrule[0.5pt]
         SAM-TS-B &12.5M &102.2M &73.39 &81.34 \\
         SAM-TS-L &30.7M &338.9M &\underline{78.37} &\underline{84.99} \\
         SAM-TS-H &58.0M &695.0M &\textbf{79.27} &\textbf{85.63} \\
         \midrule[0.5pt]
         SAM-TS-B\dag &12.5M &102.2M &85.80 &92.21 \\
         SAM-TS-L\dag &30.7M &338.9M &89.04 &94.26 \\
         SAM-TS-H\dag &58.0M &695.0M &88.59 &93.79 \\
         \bottomrule[1pt]
    \end{tabular}}
    \vspace{-3mm}
\end{table}

\begin{table*}[t!]
    \centering
    \caption{
    \textbf{Label quality comparison on COCO\_TS.} While the original labels of COCO\_TS are obtained using a weakly supervised segmentation model as described in Sec.~\ref{subsec: datasets}, we use SAM-TS-L trained on different datasets to automatically generate its labels for comparison. `COCO\_TS $\xrightarrow[]{}$ Target' means train the model on COCO\_TS and evaluate it on Target (\eg, Total-Text, TextSeg, and HierText) directly, without fine-tuning.
    }
    \label{tab: cocots}
    \setlength{\tabcolsep}{6pt}
    \resizebox{\linewidth}{!}{
    \begin{tabular}{l|cccc|cccc|cccc}
         \toprule[1pt]
         \rowcolor{gray!20} &\multicolumn{4}{c|}{\textbf{COCO\_TS $\xrightarrow[]{}$ Total-Text}} &\multicolumn{4}{c|}{\textbf{COCO\_TS $\xrightarrow[]{}$ TextSeg}} &\multicolumn{4}{c}{\textbf{COCO\_TS $\xrightarrow[]{}$ HierText}} \\
         \cline{2-5} \cline{6-9} \cline{10-13}
         \rowcolor{gray!20} \multirow{-2}{*}{\textbf{Model for Auto-labeling COCO\_TS}} &\textbf{fgIOU} &\textbf{$\Delta$} &\textbf{F-score} &\textbf{$\Delta$} &\textbf{fgIOU} &\textbf{$\Delta$} &\textbf{F-score} &\textbf{$\Delta$} &\textbf{fgIOU} &\textbf{$\Delta$} &\textbf{F-score} &\textbf{$\Delta$} \\
         \midrule[1pt]
         Original~\cite{bonechi2019coco_ts} &65.45 &$-$ &75.84 &$-$ &58.54 &$-$ &84.48 &$-$ &53.83 &$-$ &66.21 &$-$ \\
         \midrule[0.5pt]
         SAM-TS-L trained on Total-Text &80.02 &\textcolor[rgb]{0,0.8,0}{+14.57} &87.33 &\textcolor[rgb]{0,0.8,0}{+11.49} &56.93 &\textcolor{red}{-1.61} &84.13 &\textcolor{red}{-0.35} &63.64 &\textcolor[rgb]{0,0.8,0}{+9.81} &71.63 &\textcolor[rgb]{0,0.8,0}{+5.42} \\
         SAM-TS-L trained on TextSeg &76.09 &\textcolor[rgb]{0,0.8,0}{+10.64} &84.10 &\textcolor[rgb]{0,0.8,0}{+8.26} &81.12 &\textcolor[rgb]{0,0.8,0}{+22.58} &91.51 &\textcolor[rgb]{0,0.8,0}{+7.03} &54.75 &\textcolor[rgb]{0,0.8,0}{+0.92} &61.26 &\textcolor{red}{-4.95} \\
         SAM-TS-L trained on HierText &76.23 &\textcolor[rgb]{0,0.8,0}{+10.78} &86.33 &\textcolor[rgb]{0,0.8,0}{+10.49} &71.23 &\textcolor[rgb]{0,0.8,0}{+12.69} &86.87 &\textcolor[rgb]{0,0.8,0}{+2.39} &69.95 &\textcolor[rgb]{0,0.8,0}{+16.12} &79.67 &\textcolor[rgb]{0,0.8,0}{+13.46} \\
         \bottomrule[1pt]
    \end{tabular}}
\end{table*}

%---------------------------------
\begin{figure*}[!t]
    \centering
    \includegraphics[width=1\linewidth]{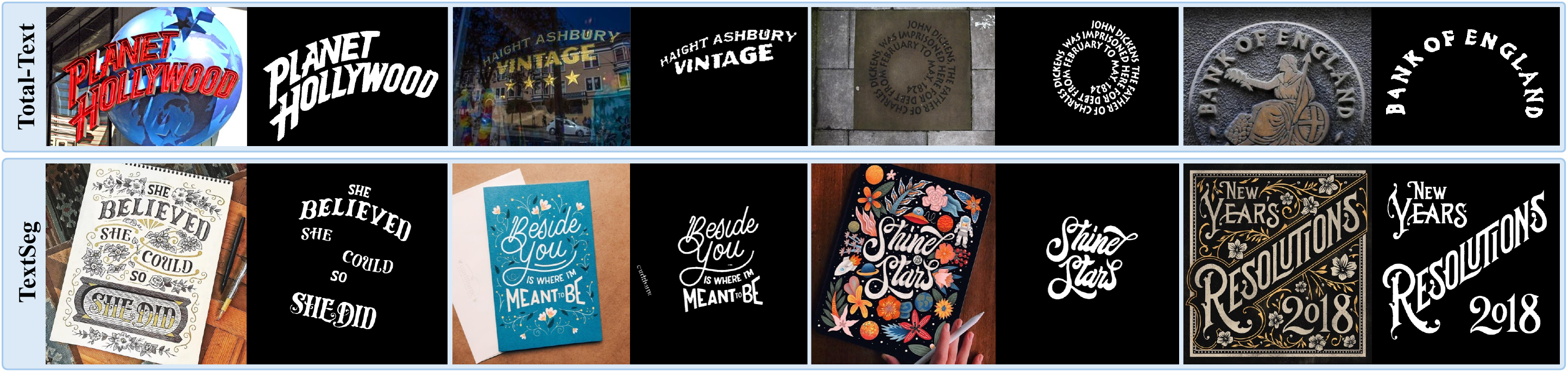}
    \caption{\textbf{Qualitative results on Total-Text and TextSeg.} The predictions are made by SAM-TS-L.}
    \label{fig: vis_tt_textseg}
    \vspace{-3mm}
\end{figure*}
%---------------------------------

%---------------------------------
\begin{figure}[!t]
    \centering
    \includegraphics[width=1\linewidth]{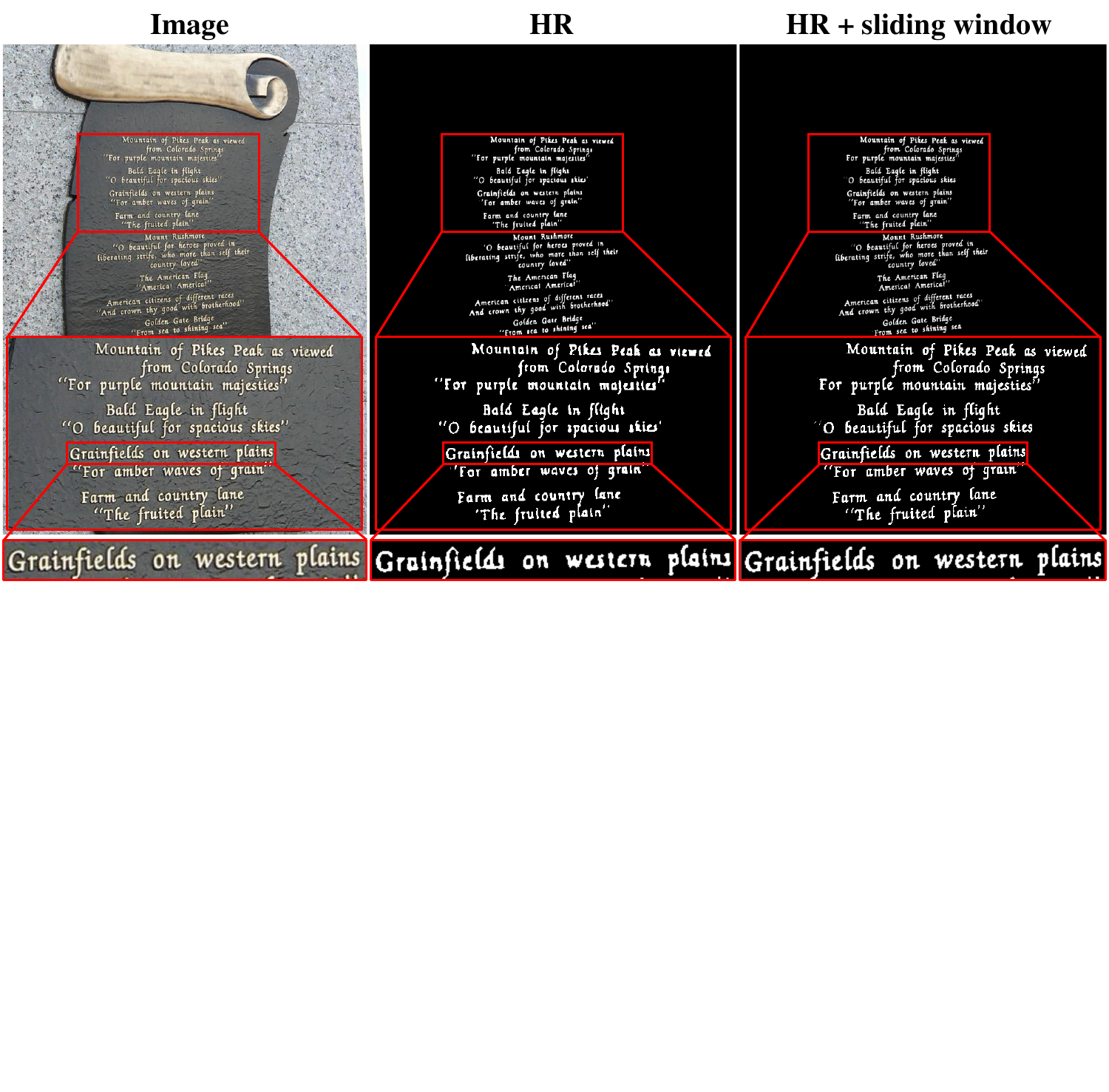}
    \caption{\textbf{Visualizations on HierText}. `HR': predictions from high-resolution mask feature. Using the sliding window strategy substantially improves the segmentation quality on extremely small texts. For example, in the bottom row of sliced patches, the tiny hollows in character `e' are distinguished when using sliding window. And the pixel-level text segmentation delivers more correct and clearer structure details. Best view on screen with zooming in.}
    \label{fig: vis_hiertext}
\end{figure}
%---------------------------------

\noindent\textbf{Influence of the tuning method and pretrained weights for initialization.}
In Tab.~\ref{tab: influence of SAM on Hi-SAM}, we further analyze the impact of pretrained weights on Hi-SAM. It is evident that adapter tuning outperforms model tuning. Using pretrained weights from SAM results in significant improvements compared to solely adopting the MAE pretrained ViT.

\subsection{Comparison with State-of-the-art Methods}
\subsubsection{Pixel-level Text Segmentation}
\noindent\textbf{Total-Text Benchmark.}
Compared to other methods, our approach stands out as the first to leverage the vision foundation model~\cite{kirillov2023segment} and realizes high-resolution pixel-level text segmentation. Although previous methods resort to utilizing additional text recognizers to enhance the semantic features~\cite{xu2022bts}, or extra text detectors and annotations~\cite{yu2023scene}, SAM-TS still outperforms them and achieves the best fgIOU performance on Total-Text. As shown in Tab.~\ref{tab: comp on totaltext}, compared to TFT~\cite{yu2023scene}, SAM-TS-L achieves a gain of 2.49\% fgIOU and SAM-TS-H presents state-of-the-art 84.86\% fgIOU.
Compared to PGTSNet~\cite{xu2022bts}, SAM-TS-L significantly outperforms it by 5.49\% fgIOU.

\noindent\textbf{TextSeg Benchmark.}
As shown in Tab.~\ref{tab: comp on textseg}, our method delivers commendable performance on TextSeg. SAM-TS-B achieves 87.15\% fgIOU and SAM-TS-L further improves the fgIOU performance by 1.62\%. Moreover, SAM-TS-H reaches 88.96\% fgIOU. Some visualization examples on Total-Text and TextSeg are shown in Fig.~\ref{fig: vis_tt_textseg}.

\begin{table*}[t!]
    \centering
    \caption{
    \textbf{Comparison between Unified Detector~\cite{long2022towards} and Hi-SAM.} `TL': text-line. `WG': word grouping. `Para.': paragraph. `VFM': whether the vision foundation model is used. As for the model parameter statistics,  Unified Detector has 88.2M trainable and also total parameters. Hi-SAM-B has 16.7M trainable parameters and 106.4M parameters in total. Hi-SAM-L owns 34.9M trainable parameters and 343.1M parameters in total. Hi-SAM-H has 62.2M trainable parameters and 699.2M parameters in total.
    }
    \label{tab: comp with UD}
    \resizebox{\linewidth}{!}{
    \begin{tabular}{l|c|c|c|c|c|c|c|c|c}
         \toprule[1pt]
         \rowcolor{gray!20} \textbf{Method} &\textbf{Pixel-level Text} &\textbf{Word} &\textbf{TL} &\textbf{WG in TL} &\textbf{Para.} &\textbf{Layout (WG in Para.)} &\textbf{Promptable} &\textbf{Training on HierText} &\textbf{VFM}\\
         \midrule[0.5pt]
         Unified Detector~\cite{long2022towards} &\xmark &\tmark &\xmark &\tmark &\xmark &\tmark &\xmark &128 TPUv3 cores, $\sim$3091 epochs &No\\
         Hi-SAM (Ours) &\tmark &\tmark &\tmark &\tmark &\tmark &\tmark &\tmark &8 V100 GPUs, 150 epochs &Yes\\
         \bottomrule[1pt]
    \end{tabular}}
\end{table*}

\noindent\textbf{HierText Benchmark.}
HierText presents a significant challenge in segmenting text with fine details on both natural and document scenarios.
We evaluate TextRNet~\cite{xu2021rethinking} with the HRNetV2-W48~\cite{wang2020deep} backbone on HierText for comparison. As can be seen in Tab.~\ref{tab: comp on hiertext}, TextRNet only achieves 55.50\% fgIOU, running at 9.3 FPS tested on one V100 GPU with one batch size. In comparison, SAM-TS-B achieves 73.39\% fgIOU with 6.4 FPS. SAM-TS-L achieves 78.37\% fgIOU with 2.8 FPS and SAM-TS-H achieves 79.27\% fgIOU with 1.6 FPS. Different from Total-Text and TextSeg, segmenting texts in high-quality for HierText requires larger feature size. With the shorter side of image resized to 2,048, TextRNet achieves 70.77\% fgIOU with 2.2 FPS. We apply the sliding window strategy and the fgIOU performance of SAM-TS-L is further improved by 10.67\%, outperforming TextRNet by 18.27\%.
We provide some qualitative comparison with and without the sliding window strategy in Fig.~\ref{fig: vis_hiertext}. Using the sliding window strategy substantially improves the segmentation quality on extremely small texts.

\begin{table*}[t!]
    \centering
    \caption{
    \textbf{Comparison results on HierText test set.} fgIOU, PQ, and F-score are the primary metrics.
    }
    \label{tab: test set hiertext hidet}
    \setlength{\tabcolsep}{2pt}
    \resizebox{\linewidth}{!}{
    \begin{tabular}{l|cc|ccccc|ccccc|ccccc}
         \toprule[1pt]
         \rowcolor{gray!20} &\multicolumn{2}{c|}{\textbf{Pixel-level Text}} &\multicolumn{5}{c|}{\textbf{Word}} &\multicolumn{5}{c|}{\textbf{Text-line}} &\multicolumn{5}{c}{\textbf{Layout Analysis}} \\
         \cline{2-3} \cline{4-8} \cline{9-13} \cline{14-18}
         \rowcolor{gray!20} \multirow{-2}{*}{\textbf{Method}} &\textbf{fgIOU} &\textbf{F-score} &\textbf{PQ} &\textbf{F-score} &P &R &T &\textbf{PQ} &\textbf{F-score} &P &R &T &\textbf{PQ} &\textbf{F-score} &P &R &T \\
         \midrule[1pt]
         Unified Detector~\cite{long2022towards} &$-$ &$-$ &48.21 &61.51 &67.54 &56.47 &\textbf{78.38} &62.23 &79.91 &79.64 &\underline{80.19} &77.87 &53.60 &68.58 &76.04 &62.45 &\textbf{78.17} \\
         \midrule[0.5pt]
         Hi-SAM-B &70.94 &79.78 &59.74 &78.34 &83.97 &73.41 &76.25 &63.34 &82.15 &89.17 &76.16 &77.10 &54.46 &71.15 &78.53 &65.03 &76.54 \\
         Hi-SAM-L &\underline{75.22} &\underline{82.90} &\underline{63.10} &\underline{81.83} &\underline{87.22} &\underline{77.06} &77.11 &\underline{66.17} &\underline{84.85} &\underline{90.66} &79.74 &\underline{77.99} &\underline{57.61} &\underline{74.49} &\underline{80.43} &\underline{69.37} &77.34 \\
         Hi-SAM-H &\textbf{75.73} &\textbf{83.36} &\textbf{64.30} &\textbf{82.86} &\textbf{87.66} &\textbf{78.56} &\underline{77.60} &\textbf{66.96} &\textbf{85.30} &\textbf{91.09} &\textbf{80.20} &\textbf{78.50} &\textbf{59.09} &\textbf{75.97} &\textbf{81.52} &\textbf{71.13} &\underline{77.79} \\
         \bottomrule[1pt]
    \end{tabular}}
\end{table*}

\begin{table*}[t!]
    \centering
    \caption{
    \textbf{Comparison results on HierText validation set.} `\ddag' indicates that we calculate centroid points of the polygon labels of legible words and adopt them as prompts during inference.
    }
    \label{tab: val set hiertext hidet}
    \setlength{\tabcolsep}{2pt}
    \resizebox{\linewidth}{!}{
    \begin{tabular}{l|cc|ccccc|ccccc|ccccc}
         \toprule[1pt]
         \rowcolor{gray!20} &\multicolumn{2}{c|}{\textbf{Pixel-level Text}} &\multicolumn{5}{c|}{\textbf{Word}} &\multicolumn{5}{c|}{\textbf{Text-line}} &\multicolumn{5}{c}{\textbf{Layout Analysis}} \\
         \cline{2-3} \cline{4-8} \cline{9-13} \cline{14-18}
         \rowcolor{gray!20} \multirow{-2}{*}{\textbf{Method}} &\textbf{fgIOU} &\textbf{F-score} &\textbf{PQ} &\textbf{F-score} &P &R &T &\textbf{PQ} &\textbf{F-score} &P &R &T &\textbf{PQ} &\textbf{F-score} &P &R &T \\
         \midrule[1pt]
         Unified Detector~\cite{long2022towards} &$-$ &$-$ &48.47 &61.65 &67.18 &56.96 &\textbf{78.47} &61.29 &78.66 &78.98 &78.34 &77.92 &52.86 &67.73 &76.17 &60.98 &\underline{78.04} \\
         \midrule[0.5pt]
         Hi-SAM-B &70.35 &79.65 &59.20 &77.48 &83.51 &72.26 &76.41 &62.89 &81.44 &88.89 &75.13 &77.22 &55.66 &72.64 &80.06 &66.48 &76.63 \\
         Hi-SAM-L &\underline{74.86} &\underline{82.87} &\underline{62.60} &\underline{80.99} &\underline{86.48} &\underline{76.16} &77.29 &\underline{65.62} &\underline{84.04} &\underline{90.29} &\underline{78.59} &\underline{78.08} &\underline{58.83} &\underline{75.84} &\underline{81.72} &\underline{70.75} &77.57 \\
         Hi-SAM-H &\textbf{75.45} &\textbf{83.21} &\textbf{63.83} &\textbf{82.08} &\textbf{87.02} &\textbf{77.67} &\underline{77.76} &\textbf{66.70} &\textbf{84.82} &\textbf{90.97} &\textbf{79.45} &\textbf{78.64} &\textbf{60.34} &\textbf{77.22} &\textbf{83.47} &\textbf{71.84} &\textbf{78.14} \\
         \midrule[0.5pt]
         Hi-SAM-B\ddag &70.35 &79.65 &61.17 &80.55 &84.35 &77.08 &75.93 &67.32 &88.17 &91.21 &85.32 &76.35 &57.85 &75.88 &81.43 &71.03 &76.24 \\
         Hi-SAM-L\ddag &74.86 &82.87 &64.63 &84.08 &87.79 &80.68 &76.87 &69.58 &90.06 &92.28 &87.94 &77.27 &60.42 &78.16 &82.35 &74.37 &77.30 \\
         Hi-SAM-H\ddag &75.45 &83.21 &65.73 &85.06 &88.18 &82.15 &77.28 &70.61 &90.81 &92.85 &88.85 &77.76 &61.81 &79.40 &83.86 &75.38 &77.85 \\
         \bottomrule[1pt]
    \end{tabular}}
\end{table*}

\noindent\textbf{Auto-labeling on COCO\_TS.}
We further compare the quality of labels generated by our SAM-TS-L and the original labels on COCO\_TS. We train three SAM-TS-L models on Total-Text, TextSeg, and HierText to automatically label the images in COCO\_TS using the sliding window strategy (window size: 512 $\times$ 512, stride: 384). Then, we train SAM-TTS using these different labels on COCO\_TS for 40 epochs and evaluate them on other datasets without fine-tuning. 
The ability to generalize performance across various datasets serves as a reliable indicator of the quality of labels.
Note that the original labels of COCO\_TS contain a lot of uncertain regions, we involve them for training the baseline model following previous methods~\cite{bonechi2019coco_ts,xu2021rethinking}.

The results are reported in Tab.~\ref{tab: cocots}. As can be seen, the models trained with our generated labels show significantly better generalization performance in general, which indicates that our model has an excellent auto-labeling ability. 
Note that employing labels obtained from the model trained on Total-Text does not result in improved performance on TextSeg. This is likely due to the fact that TextSeg encompasses both scenes and designed texts, whereas Total-Text predominantly emphasizes scene texts.

\subsubsection{Hierarchical Text Segmentation and Layout Analysis}
We evaluate the hierarchical text segmentation and layout analysis performance of Hi-SAM on HierText.
We first compare the functionality and training cost between the SOTA Unified Detector (UD)~\cite{long2022towards} and our Hi-SAM.

\noindent\textbf{Comparison with UD.}
UD uses the MaX-DeepLab-S backbone~\cite{wang2021max}, which has a relatively similar trainable parameter amount as Hi-SAM-H. The input image size of UD is the same as our Hi-SAM. Each object query in UD is trained to segment words in one text-line as opposed to a single word. Thus, as shown in Tab.~\ref{tab: comp with UD}, UD can perform word grouping at the text-line level naturally and also word instance segmentation with post-processing. UD leverages a layout branch to cluster object queries into different groups, thereby achieving layout analysis (word grouping at paragraph level). However, UD is not promptable and cannot segment pixel-level text masks and achieve contact text-line and paragraph masks. Besides, UD is trained on 128 TPUv3 cores with a batch size of 256 for 100K steps (about 3,091 epochs). In comparison, Hi-SAM is the first text-centric model that can perform all tasks mentioned above and is promptable. Moreover, Hi-SAM requires $20\times$ fewer training epochs and is trained on eight NVIDIA V100 GPUs.

%---------------------------------
\begin{figure*}[!t]
    \centering
    \includegraphics[width=1\linewidth]{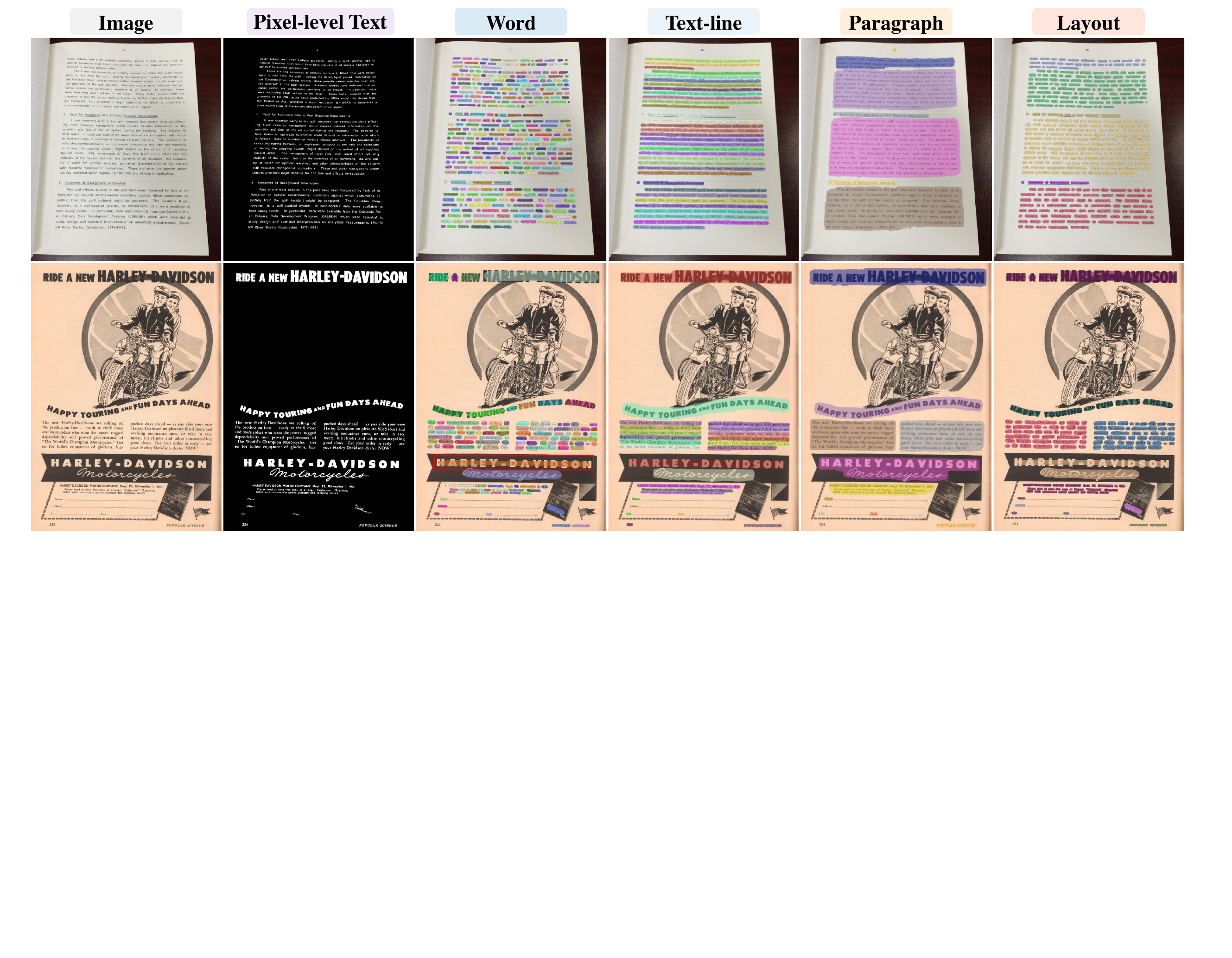}
    \caption{\textbf{Visualizations of hierarchical text segmentation and layout analysis results.} 
    Some flaws are marked with \textcolor{red}{red} boxes.
    }
    \label{fig: vis_hierseg}
    \vspace{-3mm}
\end{figure*}
%---------------------------------
\begin{figure}[!t]
    \centering
    \includegraphics[width=1\linewidth]{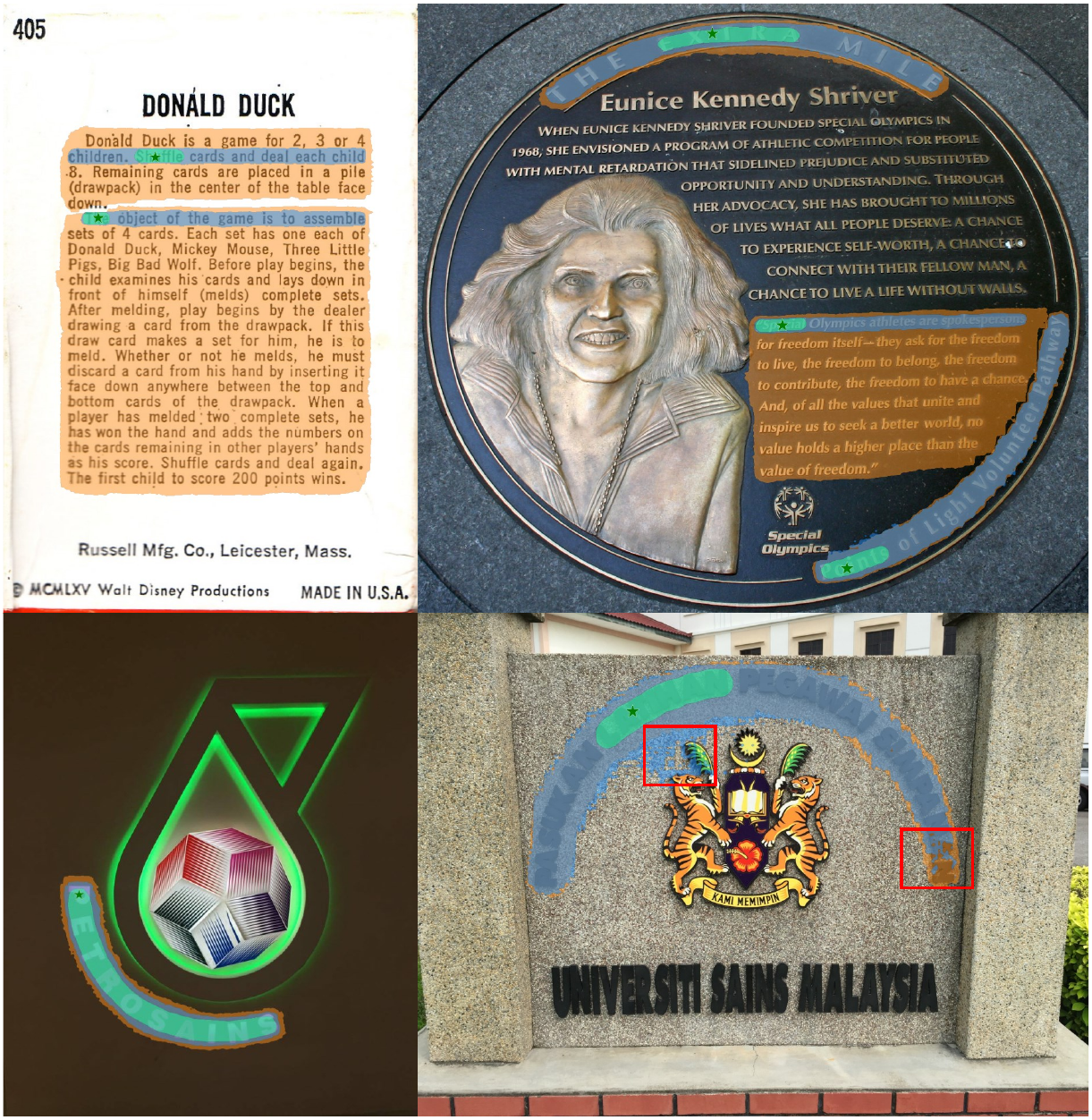}
    \caption{\textbf{Visualizations of promptable segmentation.} Green stars in images are point prompt locations. Word, text-line, and paragraph masks are in \textcolor{green}{green}, \textcolor{blue}{blue}, and \textcolor{orange}{orange} color.}
    \label{fig: vis_prompt_seg}
    \vspace{-3mm}
\end{figure}
%---------------------------------

\noindent\textbf{Results on Test Set.}
As presented in Tab.~\ref{tab: test set hiertext hidet}, compared to UD, our method achieves significant improvements in terms of PQ and F-score for word level, text-line level, and layout analysis. 
For example, Hi-SAM-H outperforms UD by 16.09\% PQ and 21.35\% F-score at word level, 4.73\% PQ and 5.39\% F-score at text-line level word grouping, 5.49\% PQ and 7.39\% F-score at paragraph level layout analysis. We notice that UD performs poorly at the word level, probably because UD cannot separate adjacent instances better in dense text scenarios. Whereas, this does not affect the performance at the text-line and paragraph level. Note that UD utilizes all samples in each image for training and performs layout analysis by calculating affinity scores among object queries. In contrast, we design completely different training and inference pipelines, where we only sample a few samples in each image for training and implement layout analysis without an additional layout branch. Even with 20$\times$ fewer training epochs, our method still achieves significantly better performance.
Comparing different Hi-SAM variants, we observe that Hi-SAM-L is significantly better than Hi-SAM-B, \ie, achieving improvement by 3.36\% PQ and 3.49\% F-score at the word level, by 2.83\% PQ and 2.70\% F-score on text-line level, and by 3.15\% PQ and 3.34\% F-score at the paragraph level.
Comparing Hi-SAM-H to Hi-SAM-L, the main improvement lies in the layout analysis results, owing to the better image embedding.

\noindent\textbf{Results on Validation Set.}
We also report the results on the HierText validation set, as shown in Tab.~\ref{tab: val set hiertext hidet}. The results of UD are collected from the official GitHub repository\footnote{\url{https://github.com/google-research-datasets/hiertext}}.
As can be seen, compared to UD, our method also achieves similar performance improvements as on the test set. Hi-SAM-H outperforms UD by 15.36\% PQ and 20.43\% F-score at word level, 5.41\% PQ and 6.16\% F-score at text-line level word grouping, 7.48\% PQ and 9.49\% F-score at paragraph level layout analysis.
In addition, we use the polygon annotation of legible words to calculate centroid points and adopt these points as prompts for generating hierarchical masks during inference. We observe that using these centroid points as prompts mainly enhances the recall (R) scores, especially at the text-line level. For instance, for Hi-SAM-B, the recall scores are improved by 4.82\%, 10.19\%, and 4.55\% at the word level, text-line level, and layout analysis, respectively. For Hi-SAM-H, the recall metrics are enhanced by 4.49\%, 9.40\%, and 3.54\% at the word level, text-line level, and layout analysis, respectively. It indicates that there is still room for developing a better sampling strategy. Perhaps leveraging a simple counting module for providing a text center heat map may help. In addition, since we apply NMS on text-line masks and discard corresponding word-level masks, the remained word-level masks may not achieve the best segmentation.

\begin{table}[t!]
\centering
\caption{
\textbf{AP comparison on HierText validation set.}
}
\label{tab: AP}
\setlength{\tabcolsep}{3pt}
\resizebox{\linewidth}{!}{
\begin{tabular}{l|ccc|ccc}
     \toprule[1pt]
     \rowcolor{gray!20} &\multicolumn{3}{c|}{\textbf{Text-line}} &\multicolumn{3}{c}{\textbf{Layout Analysis}} \\
     \cline{2-4} \cline{5-7}
     \rowcolor{gray!20} \multirow{-2}{*}{\textbf{Method}} &\textbf{AP} &\textbf{AP$_{50}$} &\textbf{AP$_{75}$} &\textbf{AP} &\textbf{AP$_{50}$} &\textbf{AP$_{75}$} \\
     \midrule[1pt]
     Unified Detector~\cite{long2022towards} &40.2 &73.6 &40.5 &22.3 &41.3 &22.9 \\
     \midrule[0.5pt]
     Hi-SAM-B &40.0 &73.3 &41.9 &25.2 &48.3 &24.9 \\
     Hi-SAM-L &\underline{43.1} &\underline{76.2} &\underline{46.4} &\underline{28.5} &\underline{52.5} &\underline{29.6} \\
     Hi-SAM-H &\textbf{44.7} &\textbf{77.3} &\textbf{50.0} &\textbf{29.4} &\textbf{53.2} &\textbf{32.1} \\
     \midrule[1pt]
\end{tabular}}
\vspace{-5mm}
\end{table}

On the validation set, we compared the AP metric at IoU thresholds ranging from 0.5 to 0.95, as shown in Tab.~\ref{tab: AP}. At the text-line level, the Hi-SAM models significantly outperform the UD model, particularly at higher IoU thresholds. For instance, Hi-SAM-B, Hi-SAM-L, and Hi-SAM-H achieve AP$_{75}$ improvements of 1.4\%, 5.9\%, and 9.5\%, respectively. Moreover, Hi-SAM shows even greater improvements over UD in layout analysis compared to the text-line level. Specifically, Hi-SAM-L outperforms UD by 2.9\% AP on the text-line level and by 6.2\% AP in layout analysis, indicating Hi-SAM's superior segmentation accuracy, especially at the paragraph level. Among the Hi-SAM models, Hi-SAM-L notably outperforms Hi-SAM-B, with a significantly better AP$_{75}$, demonstrating the advantages of a larger backbone.

\noindent\textbf{Visualizations.}
We provide some visualizations of automatic mask generation and promptable segmentation in Fig.~\ref{fig: vis_hierseg} and Fig.~\ref{fig: vis_prompt_seg}. Some failure cases are marked with red boxes.
For example, the word segments for long words are sometimes incomplete. The text-line segmentation results in extremely curved entities are not promising, which is caused by the quadrilateral labels at the text-line level.

\section{Limitation and Discussion}
\label{sec:limitation_discussion}
While Hi-SAM has shown promising performance, there is still potential for further improvement. 1) Due to the heavy backbone, the existing Hi-SAM models cannot reach real-time inference speed. As some follow-ups of SAM~\cite{zhao2023fast,zhang2023faster} have achieved great progress in developing lightweight SAM models, it is promising to develop efficient Hi-SAM models by exploring a more lightweight structure.
2) Since we freeze the pretrained weights of SAM's image encoder and only use HierText to train Hi-SAM, the generalization ability of Hi-SAM for unseen domains can be constrained.
3) Hi-SAM requires labels across four text hierarchies, which are scarce in real datasets. It is worth trying to explore synthetic or diffusion-based~\cite{ho2020denoising} methods to generate large-scale hierarchical text data with complex layouts and accurate text contents.
4) We observe that the pixel-level text segmentation results of Hi-SAM in Tab.~\ref{tab: test set hiertext hidet} are lower than those in Tab.~\ref{tab: comp on hiertext}. How to better synergize different text hierarchies during training and inference is also an open issue.
5) As a text-centric model, Hi-SAM falls short in conducting the multi-class (including text, title, list, table, and figure) layout analysis in some document-only datasets, such as PubLayNet~\cite{zhong2019publaynet}.

\section{Conclusion}
\label{sec:conclusion}
In this study, we demonstrate the transformation of SAM into a cutting-edge pixel-level text segmentation model, SAM-TS, through minimal customizations. It is capable of generating high-quality labels semi-automatically, helping unify annotations across four text hierarchies in HierText. Building upon these advancements, we introduce Hi-SAM -- a pioneering unified model for hierarchical text segmentation. It seamlessly operates across multiple levels, spanning from the finest pixel level to paragraph, while conducting layout analysis without the need for any dedicated modules. 
Extensive experimental results unequivocally highlight the superior performance of our proposed method compared to existing representative approaches. We hope this work can pave the way for practical hierarchical text segmentation in real-world text images and inspire further exploration in this field.

{
\bibliographystyle{IEEEtran}
 \bibliography{egbib}
}

\end{document}